\pdfoutput=1

\documentclass[11pt,table,xcdraw]{article}
\usepackage[final]{acl}
\usepackage{times}
\usepackage{latexsym}
\usepackage[T1]{fontenc}
\usepackage[utf8]{inputenc}
\usepackage{microtype}
\usepackage{inconsolata}

\usepackage{xspace}
\usepackage{multirow}
\usepackage{booktabs}
\usepackage{bm}
\usepackage{amsmath}
\usepackage{amsfonts}
\usepackage{hyperref}
\usepackage{graphicx}
\usepackage{framed}
\usepackage{enumitem}
\usepackage{pifont}
\usepackage{CJKutf8}
\usepackage{marvosym}
\usepackage{lipsum}
\usepackage{pmboxdraw}
\usepackage{caption}
\usepackage{subcaption}
\usepackage{epigraph}
\usepackage{longtable}
\usepackage{tabularx}
\usepackage{arydshln}
\usepackage{colortbl}

\newcommand{\eg}{\emph{e.g.,}\xspace}

\newcommand{\ignore}[1]{}

\newcommand{\mmath}{{MMATH}\xspace}

\newcommand{\encot}{EN-CoT\xspace}

\newcommand{\nativethink}{Native-Think\xspace}
\newcommand{\enthink}{EN-Think\xspace}
\newcommand{\ensft}{EN-SFT\xspace}

\newcommand{\atp}{ATP\xspace}
\newcommand{\dit}{DIT\xspace}
\newcommand{\qrt}{QRT\xspace}

\makeatletter
\def\@fnsymbol#1{}
\makeatother

\title{\mmath: A Multilingual Benchmark for Mathematical Reasoning}

\author{
\textbf{Wenyang Luo}\textsuperscript{\rm{1}}, 
\textbf{Wayne Xin Zhao}\textsuperscript{\rm{1 \Letter}\thanks{\textsuperscript{\Letter}\ Corresponding author}\ }\textbf{,} \\
\textbf{Jing Sha}\textsuperscript{\rm{2}}\textbf{,} 
\textbf{Shijin Wang}\textsuperscript{\rm{2}}\textbf{,} 
\textbf{Ji-Rong Wen}\textsuperscript{\rm{1}} \\
\textsuperscript{1} Gaoling School of Artificial Intelligence, Renmin University of China \\
\textsuperscript{2} iFLYTEK Research (Central China), iFLYTEK Co., Ltd. \\
\texttt{wenyang\_luo@outlook.com \ \  batmanfly@gmail.com \ \ \{jingsha,sjwang3\}@iflytek.com} \\
}

\begin{document}
\maketitle
\begin{abstract}
The advent of large reasoning models, such as OpenAI o1 and DeepSeek R1, has significantly advanced complex reasoning tasks. However, their capabilities in multilingual complex reasoning remain underexplored, with existing efforts largely focused on simpler tasks like MGSM. To address this gap, we introduce \textbf{\mmath}, a benchmark for multilingual complex reasoning spanning 374 high-quality math problems across 10 typologically diverse languages. Using \mmath, we observe that even advanced models like DeepSeek R1 exhibit substantial performance disparities across languages and suffer from a critical \textit{off-target} issue—generating responses in unintended languages. To address this, we explore strategies including prompting and training, demonstrating that reasoning in English and answering in target languages can simultaneously enhance performance and preserve target-language consistency. Our findings offer new insights and practical strategies for advancing the multilingual reasoning capabilities of large language models. Our code and data could be found at \url{https://github.com/RUCAIBox/MMATH}.

\end{abstract}

\section{Introduction}
Large language models (LLMs) have shown surprising reasoning ability in many areas, such as mathematical reasoning and logical reasoning, with the advancement of chain-of-thought (CoT). Recent research such as OpenAI o1~\cite{jaech-etal-2024-openai} and DeepSeek R1~\cite{guo-etal-2025-deepseek}, has further improved the ability through longer CoT with intermediate plan actions and engaging in trial and error exploration, ultimately improving their performance on complex tasks.

\begin{figure}[ht]
\centering
\includegraphics[scale=0.6]{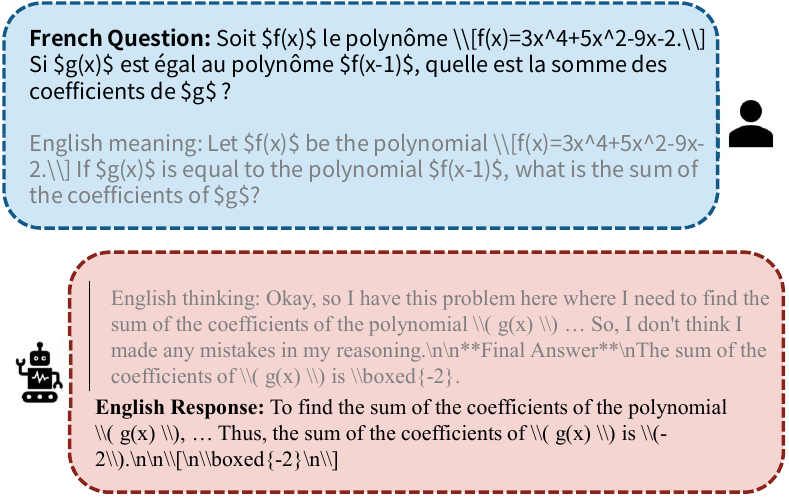}
\caption{A demonstration of off-target generation. The text with a blue background shows a French question, while the red text represents LLMs' English thinking and response, highlighting a language inconsistency.}
\label{fig:off-target demo}
\end{figure}

The imbalance in reasoning performance across different languages has drawn increasing attention since the early development~\cite{shi-etal-2022-language}. To enhance the multilingual reasoning abilities of LLMs, prior research has generally followed two main approaches: prompt-based techniques and training-based interventions. Prompt-based methods typically guide models to leverage their stronger reasoning abilities in English, while training-based methods aim to align multilingual inputs with English-centric reasoning capabilities through supervised learning~\cite{huang-etal-2023-not, zhu-etal-2024-question}. However, most of these efforts have focused on relatively \textit{simple benchmarks} such as MGSM~\cite{shi-etal-2022-language} and MSVAMP~\cite{chen-etal-2023-breaking}, while more complex benchmarks, like AIME, remain largely monolingual. This gap has limited progress in understanding and improving \textit{challenging multilingual reasoning tasks}. Moreover, the \textit{off-target issue}—where models respond in unintended languages—fails to meet the needs of monolingual users (see Figure~\ref{fig:off-target demo}), yet remains a significant but overlooked problem in previous research.

To address this gap, we introduce \textbf{\mmath}, a new benchmark specifically designed for multilingual complex reasoning. \mmath comprises 374 carefully selected math problems from high-quality sources including AIME, CNMO, and MATH-500, and covers ten typologically and geographically diverse languages. Each problem is translated and validated through a rigorous pipeline that combines frontier LLMs with human verification, ensuring semantic consistency.

Building on the \mmath benchmark, we analyze the behavior of advanced LLMs and identify a prevalent issue: off-target phenomena, where models generate responses in unintended languages. To quantify this, we introduce a metric called language consistency ratio (LCR), which measures the degree of language alignment between input and output. Our investigation centers around two key research questions:
\textbf{(1) Can LLMs solve non-English questions by reasoning in English?}
\textbf{(2) Can LLMs generate answers in the target language?}
The first question explores whether reasoning in English—a high-resource language—can enhance performance on non-English tasks, while the second addresses the practical usability of ensuring outputs are in the user's language. For the first question, we find that reasoning in English shows consistently better performance when asked in low-resource languages. For the second question, our prompting skills reveal that large models can be explicitly prompted to generate responses in the desired language, while smaller models frequently fail to retain this control. And moderate thinking intervention can greatly improve language consistency. Finally, we show that after being trained with English reasoning traces and answers in target languages, models can get substantial increases in both answer accuracy and language consistency. Qwen2.5-32B-Instruct with 3K data achieves comparable performance (66.72) with Distill-Qwen-32B (67.01), and answering LCR grows to 97.61, much more higher than reasoning models like QwQ-32B (58.94).

Our contributions are listed as follows:

$\bullet$ We propose \textbf{\mmath}, a new benchmark for evaluating multilingual complex reasoning covering 374 high-quality math problems across 10 typologically and geographically diverse languages.

$\bullet$ Our prompting techniques show that moderate thinking intervention could greatly mitigate the off-target problem, and the results may reveal more truthful ability of multilingual models.

$\bullet$ We demonstrate that training on English reasoning traces with multilingual answers could significantly improve answer accuracy and language consistency simultaneously.

\section{The \mmath benchmark}
\label{sec:dataset-mmath}
In this section, we introduce the construction process of the \mmath benchmark, a new multilingual dataset designed to evaluate complex mathematical reasoning across ten languages. We begin by describing the source data used to build the English portion of the benchmark, followed by the language selection and translation methodology.

\paragraph{Source data.}
To get high quality English mathematical reasoning benchmark, we choose the following three datasets as the data source.

$\bullet$ \textit{AIME.} American Invitational Mathematics Examination (AIME)\footnote{\url{https://maa.org/maa-invitational-competitions/}} is a challenging math contest for top high school students, requiring high logical thinking.

$\bullet$ \textit{CNMO.} China National Mathematical Olympiad (CNMO)\footnote{\url{https://www.cms.org.cn/}} is a high-level math competition in China, used to help select students for the national IMO team.

$\bullet$ \textit{MATH-500.} MATH-500~\cite{lightman-etal-2023-let} is a benchmark of 500 math problems covering topics such as algebra and calculus.



We collect 30 problems from AIME 2024, 15 from AIME 2025, 18 from CNMO, and 311 filtered problems from MATH-500, resulting in a total of 374 English examples. Most answers in \mmath are written as single LaTeX formulas or plain Arabic numerals. Since some MATH-500 problems are purely textual and may introduce bias when translated (\eg name results may have different translations), we filter them out and retain only those with numerical answers.

\paragraph{Language Selection.}
We select a total of 10 languages, including different language families. In addition to English (en), the selected languages are Chinese (zh), Arabic (ar), Spanish (es), French (fr), Japanese (ja), Korean (ko), Portuguese (pt), Thai (th), and Vietnamese (vi), resulting in a total of 3,740 examples in our \mmath benchmark.

\paragraph{Construction Process.}
To build high-quality multilingual translations of mathematical problems, we develop a three-stage pipeline that combines the strengths of large language models (LLMs) and human expertise. Our process begins with initial LLM-based translation, followed by iterative refinement through cross-model verification, and concludes with manual revision by certified human annotators. The pipeline is illustrated in Figure~\ref{fig:demo}.

$\bullet$ \emph{Stage I: Initial LLM Translation.} 
We begin by translating the mathematical problems into the target languages using a powerful large language model, such as GPT-4~\cite{achiam-etal-2023-gpt}. As for prompt design, we explicitly instruct the model to preserve all mathematical formulas unchanged and to avoid generating unnecessary text such as ``The translation is: xxx.''. Additionally, we include a one-shot example to better elicit the model’s translation capabilities. The full prompt used in this stage is provided in Table~\ref{tab:translation prompt}.

$\bullet$ \emph{Stage II: Iterative LLM Revision.} After that, we use GPT-4~\cite{achiam-etal-2023-gpt}, Claude-3.5 sonnet\footnote{\url{https://www.anthropic.com/news/claude-3-5-sonnet}} and Grok-3\footnote{\url{https://x.ai/news/grok-3}} to analyze the translation results and iteratively improve them. In one iteration, if a model approves the translation results, it will be marked as ``correct'' by that model. If not, it will replace the original translation with an improved version, and all marks already given will be removed. We repeat this for several iterations until all models agree with the translation results. The prompt is shown in Table~\ref{tab:judge prompt}. After 3 iterations, only 15\% of the translation results are changed until all models agree.

$\bullet$ \emph{Stage III: Final Human Revision.} Finally each translation undergoes manual revisions, and the evaluation details are shown in Table~\ref{tab:human evaluations}. In this stage, only 3\% of the translation results are modified.

After all stages, we get the final results and use them as our \mmath benchmark. The difference between previous work and our benchmark is shown in Table~\ref{tab:benchmark-comparison}.

\begin{table}[htbp]
\centering
\scalebox{0.7}{
\begin{tabular}{lccc}
\toprule
\textbf{Benchmark} & \textbf{\#Languages} & \textbf{\#Problems} & \textbf{Difficulty} \\
\midrule
AIME 24 & 1 & 30 & Competition level \\
AIME 25 & 1 & 15 & Competition level \\
CNMO 24 & 1 & 18 & Competition level \\
MATH-500 & 1 & 500 & Undergraduate level \\
MGSM & 10 & 250 $\times$ 10 & Grade school \\
MSVAMP & 10 & 500 $\times$ 10 & Grade school \\
\mmath (Ours) & 10 & 374 $\times$ 10 & Mixed \\
\bottomrule
\end{tabular}
}
\caption{Comparison between our \mmath and other mathematical reasoning benchmarks.
}
\label{tab:benchmark-comparison}
\end{table}

\section{Experiments}

In this section, we evaluate the multilingual reasoning abilities of popular LLMs on our \mmath benchmark.

\subsection{Experimental Setups}\label{sec:exp-setup}
\paragraph{Models.}
We conduct comprehensive evaluations on several popular models. For open-source complex reasoning models, we include QwQ-32B~\cite{qwq32b}, DeepSeek-R1~\cite{guo-etal-2025-deepseek}, and various sizes of its distilled versions. For closed-source models, we consider OpenAI's o3-mini. And for comparison, we also include chat models not specifically designed for complex reasoning, such as Gemma3-27B-IT~\cite{gemma_2025} and Qwen2.5-32B-Instruct~\cite{qwen2.5}.

\paragraph{Evaluation Prompts.}
To elicit the potential of models' reasoning ability, we prompt models with native languages~\cite{shi-etal-2022-language} and ask models to provide the final answer in a specified format (\eg within a box), enabling reliable rule-based verification of correctness. Figure \ref{fig:native-prompt} shows the native prompt of different languages.

\paragraph{Evaluation Setups.}
By default, we generate outputs using a temperature of t = 0.6, a top-p value of 0.95, and a maximum output length of 32,768 tokens. To obtain a more reliable estimate of reasoning accuracy, each evaluation is repeated 4 times, and the average result is recorded. Given the varying complexity of each benchmark subset, we report the final score using macro-average metric instead of micro-average.  We adopt two metrics answer accuracy and language consistency ratio to assess the multilingual reasoning ability of models, as defined below:

\emph{Answer Accuracy.} Answer accuracy measures the proportion of instances in which the model produces the correct final answer. To extract this answer, we employ the math extraction tool from \citet{2023opencompass}, which identifies boxed answers. If no boxed output is found, the final numerical value is extracted as a fallback. The extracted answers are then verified against the ground truth using \texttt{math\_verify}~\footnote{\url{https://github.com/huggingface/Math-Verify}}. 

\emph{Language Consistency Ratio.} The language consistency ratio (LCR) quantifies how consistently a list of detected languages matches a reference list. In our work, we use fastText~\cite{joulin-etal-2016-bag} for automatic language identification, and compute LCR to evaluate whether (question, thinking) and (question, answering) are expressed in the same language. 
We further validate its reliability by manually inspecting 100 randomly selected samples, reaching a 95\% correct ratio, which is consistent with existing work~\cite{wyawhare-etal-2023-comparative},

\begin{table*}[t]
\centering
\scalebox{0.8}{%
\begin{tabular}{lccccccccccc}
\toprule
Model & EN & ZH & AR & ES & FR & JA & KO & PT & TH & VI & AVG \\
\midrule
\multicolumn{12}{c}{Chat LLMs} \\
\cdashline{1-12}\noalign{\vskip 0.3ex}
Qwen2.5-32B-Instruct & 38.43 & 29.38 & 27.03 & 31.13 & 29.48 & 25.94 & 26.44 & 31.17 & 27.76 & 27.37 & 29.41 \\
Gemma3-27B-IT & 50.55 & 46.39 & 43.82 & 46.09 & 46.95 & 43.01 & 43.69 & 43.36 & 42.90 & 42.06 & 44.88 \\
\midrule
\multicolumn{12}{c}{Reasoning LLMs (distilled)} \\
\cdashline{1-12}\noalign{\vskip 0.3ex}
DeepSeek-R1-Distill-Qwen-1.5B & 45.41 & 37.59 & 34.50 & 40.40 & 42.08 & 35.08 & 34.40 & 35.49 & 28.06 & 36.89 & 36.99 \\
DeepSeek-R1-Distill-Qwen-7B & 63.90 & 58.53 & 56.50 & 62.81 & 61.58 & 50.90 & 59.90 & 62.72 & 48.97 & 58.53 & 58.44 \\
DeepSeek-R1-Distill-Llama-8B & 56.31 & 45.70 & 33.68 & 52.44 & 54.51 & 39.21 & 36.21 & 55.42 & 30.19 & 48.20 & 45.19 \\
DeepSeek-R1-Distill-Qwen-14B & 71.88 & 55.09 & 64.71 & 69.17 & 65.28 & 55.65 & 61.04 & 66.46 & 62.36 & 66.85 & 63.85 \\
DeepSeek-R1-Distill-Qwen-32B & 73.94 & 61.69 & 65.02 & 71.96 & 70.88 & 60.29 & 59.23 & 72.68 & 63.31 & 71.12 & 67.01 \\
\midrule
\multicolumn{12}{c}{Reasoning LLMs} \\
\cdashline{1-12}\noalign{\vskip 0.3ex}
QwQ-32B & 79.43 & 74.72 & 71.10 & 80.27 & 79.04 & 64.38 & 68.56 & 78.65 & 73.43 & 77.28 & 74.69 \\
Deepseek-R1 & 78.81& 74.03& 72.59& 79.54& 76.05& 72.69& 71.38& 79.09& 75.54& 77.43& 75.72 \\
o3-mini & 82.18& 80.95& 82.06& 79.53& 79.52& 78.21& 73.75& 83.74& 77.66& 81.37& 79.90 \\
\bottomrule
\end{tabular}
}
\caption{Evaluation results of different models on our \mmath. AVG represents the average score across languages.}
\label{tab:main_results}
\end{table*}

\subsection{Main Results}\label{sec:exp-main}
Table~\ref{tab:main_results} presents the overall results, with detailed subset-level outcomes shown in Table~\ref{tab:main_results_all}. We observe a consistent pattern of linguistic inconsistency across all benchmarks and model sizes: models perform significantly better on high-resource languages (e.g., English, Chinese) than on low-resource ones (e.g., Arabic, Thai). This disparity underscores the ongoing difficulty of achieving robust cross-lingual generalization. Interestingly, the performance gap between chat and reasoning models varies by language. In high-resource languages, reasoning models show clear advantages, while in low-resource settings, the gap narrows—suggesting that language modeling ability remains a key bottleneck.

When comparing model types, chat models perform reasonably well on simpler tasks like MATH-500 but struggle on more complex reasoning benchmarks. In contrast, reasoning models consistently outperform them, especially on harder tasks. Notably, smaller reasoning models such as DeepSeek-R1-Distill-Qwen-7B rival or surpass larger chat models, demonstrating the value of targeted reasoning supervision. Among all evaluated models, o3-mini, DeepSeek-R1, and QwQ-32B emerge as the top performers, establishing strong baselines for multilingual mathematical reasoning.

\subsection{Further Analysis}
In this section, we mainly focus on two research questions related to language consistency: \textbf{(1) Can LLMs solve non-English questions by reasoning in English?} \textbf{(2) Can LLMs generate answers in the target language?} In detail, we start by investigating the normal behavior of LLMs in Section~\ref{sec:off-target problem}. Then we introduce different methods to explicitly elicit target-language responses in Section~\ref{sec:Explicit Language Elicitation}. Finally, we train LLMs with English reasoning traces and target language answers, which proves helpful for both answer accuracy and language consistency in Section~\ref{sec:training}.

\subsubsection{The Off-target Problem of Complex Reasoning Models}\label{sec:off-target problem}

Previous study~\cite{chen-etal-2023-target} has observed the issue of off-target in some multilingual scenarios, which means the question and response language are mismatched. In this section, we investigate whether this happens in the multilingual complex reasoning area. Given that reasoning models often generate both internal thinking steps and final answers, a fundamental question emerges: What language do models exactly use during thinking and answering?

To examine this, we employ fastText as discussed in Section~\ref{sec:exp-setup} to detect the language and calculate the LCR results at both parts. We analyze three reasoning models DeepSeek-R1-Distill-Qwen-7B, DeepSeek-R1-Distill-Qwen-32B, and QwQ-32B and compare their behavior with a chat model Qwen2.5-32B-Instruct.

\paragraph{Off-Target Answering Exists.} The LCR results are shown in Table~\ref{tab:main_lcr_results}. As we can observe, though reasoning models have a high performance in accuracy as discussed in~\ref{sec:exp-main}, they have a much lower answering LCR (lower than 60\%) compared with chat models like Qwen2.5-32B-Instruct (nearly 100\%). 

\begin{table}[htbp]
\centering
\small
\scalebox{0.95}{
\begin{tabular}{lcc}
\toprule
Model & Thinking LCR & Answering LCR \\
\midrule
Distill-Qwen-7B & 45.31 & 47.11 \\
Distill-Qwen-32B & 40.12 & 45.13 \\
QwQ-32B & 57.47 & 58.94 \\
Qwen2.5-32B-Instruct & 99.51 & 99.51 \\
\bottomrule
\end{tabular}
}
\caption{Language consistency ratio for different models. Thinking LCR measures the match ratio between detected thinking language and question language; Answering LCR measures for the answer language.}
\label{tab:main_lcr_results}
\end{table}

\begin{figure}[h]
\centering
\begin{subfigure}{0.49\linewidth}
\includegraphics[width=\linewidth]{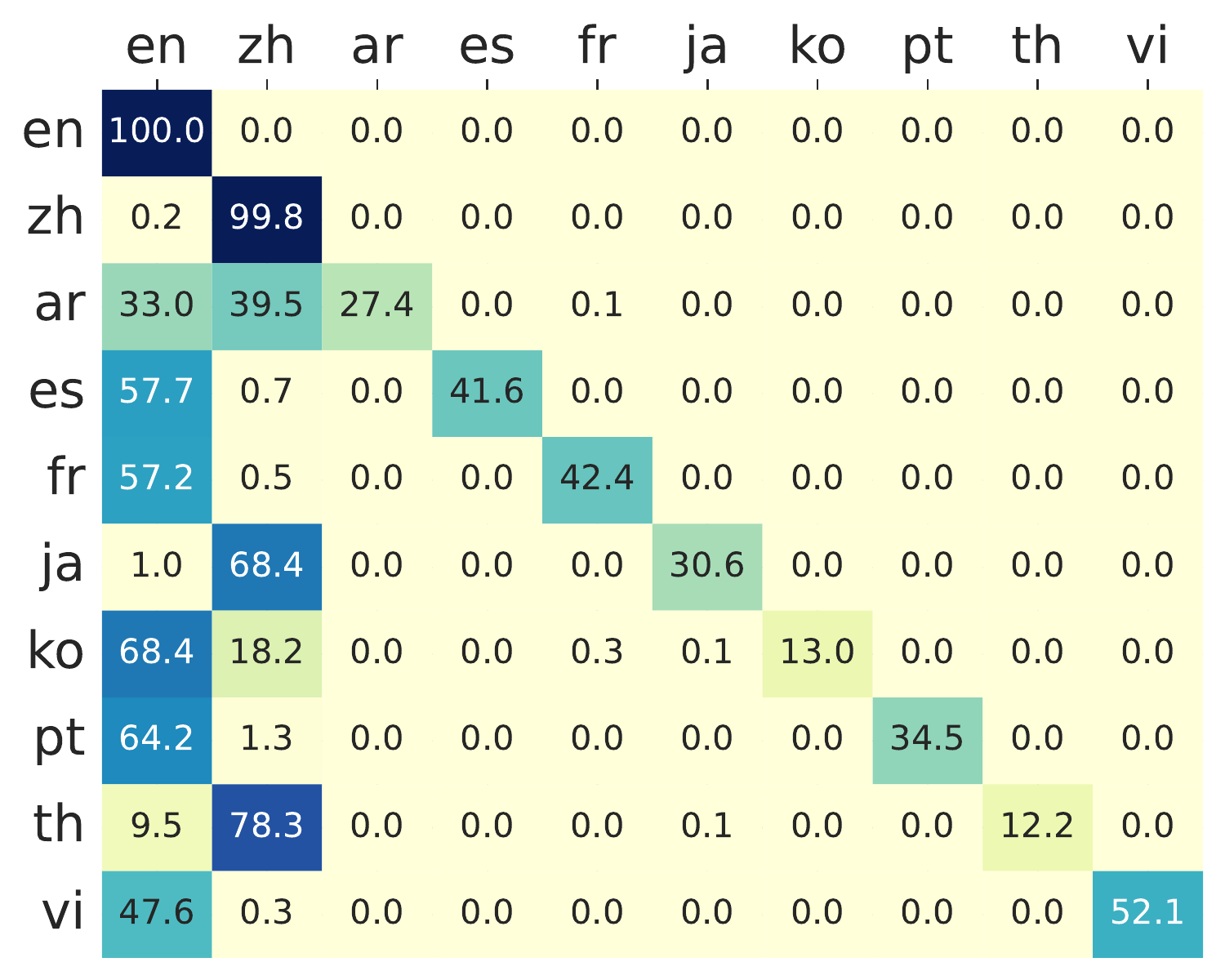}
\caption{Distill-Qwen-7B}
\end{subfigure}
\begin{subfigure}{0.49\linewidth}
\includegraphics[width=\linewidth]{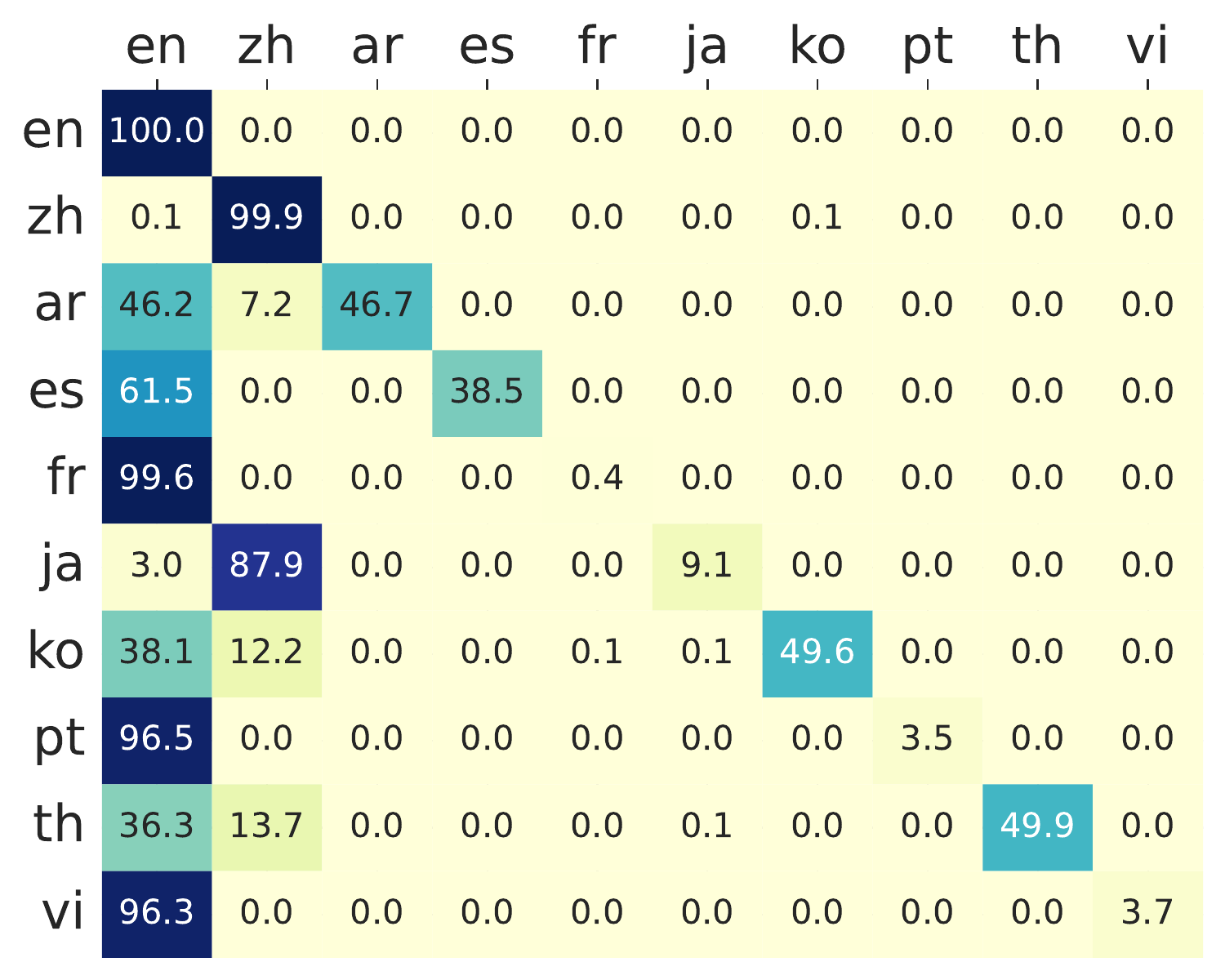}
\caption{Distill-Qwen-32B}
\end{subfigure}
\begin{subfigure}{0.49\linewidth}
\includegraphics[width=\linewidth]{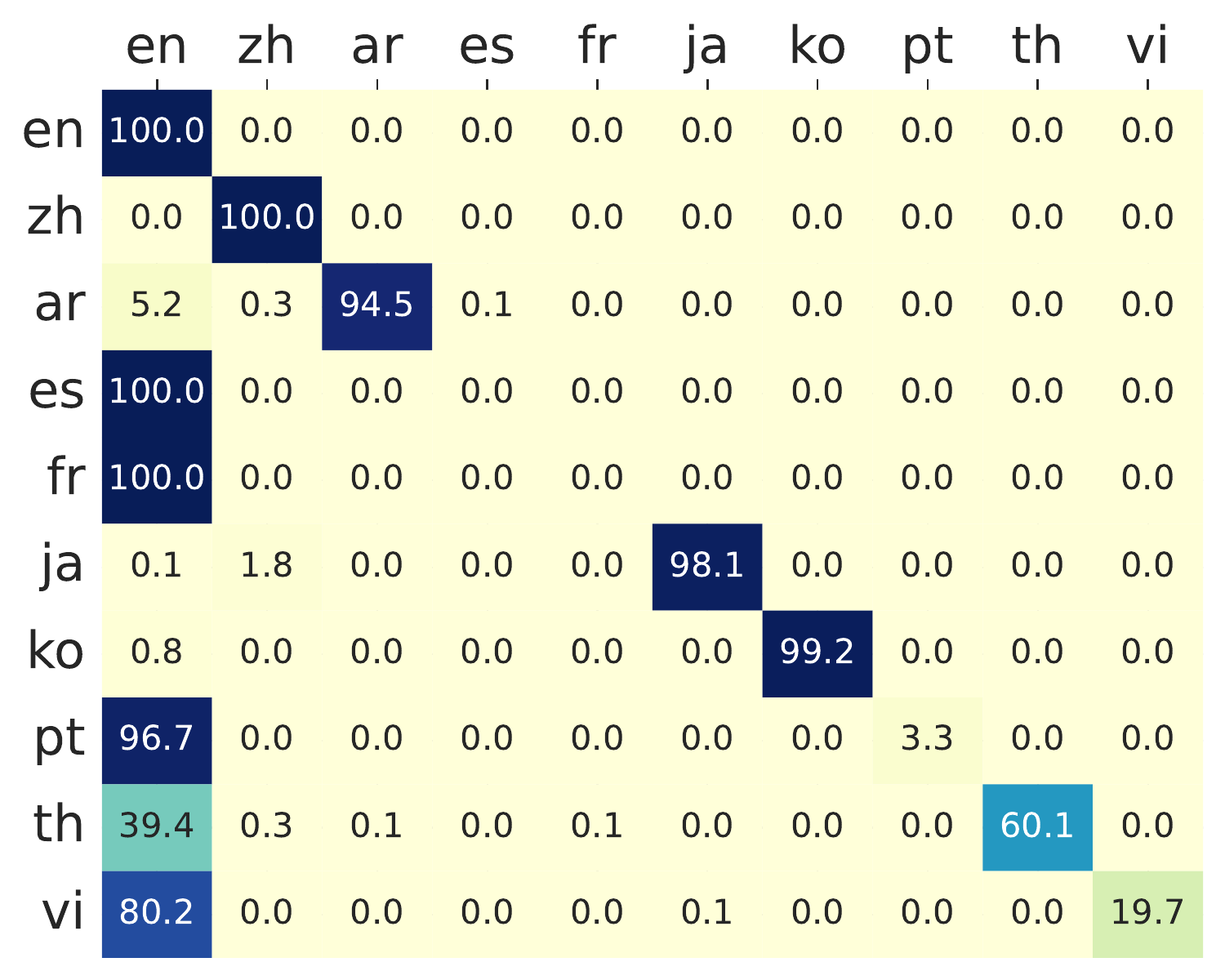}
\caption{QwQ-32B}
\end{subfigure}
\begin{subfigure}{0.49\linewidth}
\includegraphics[width=\linewidth]{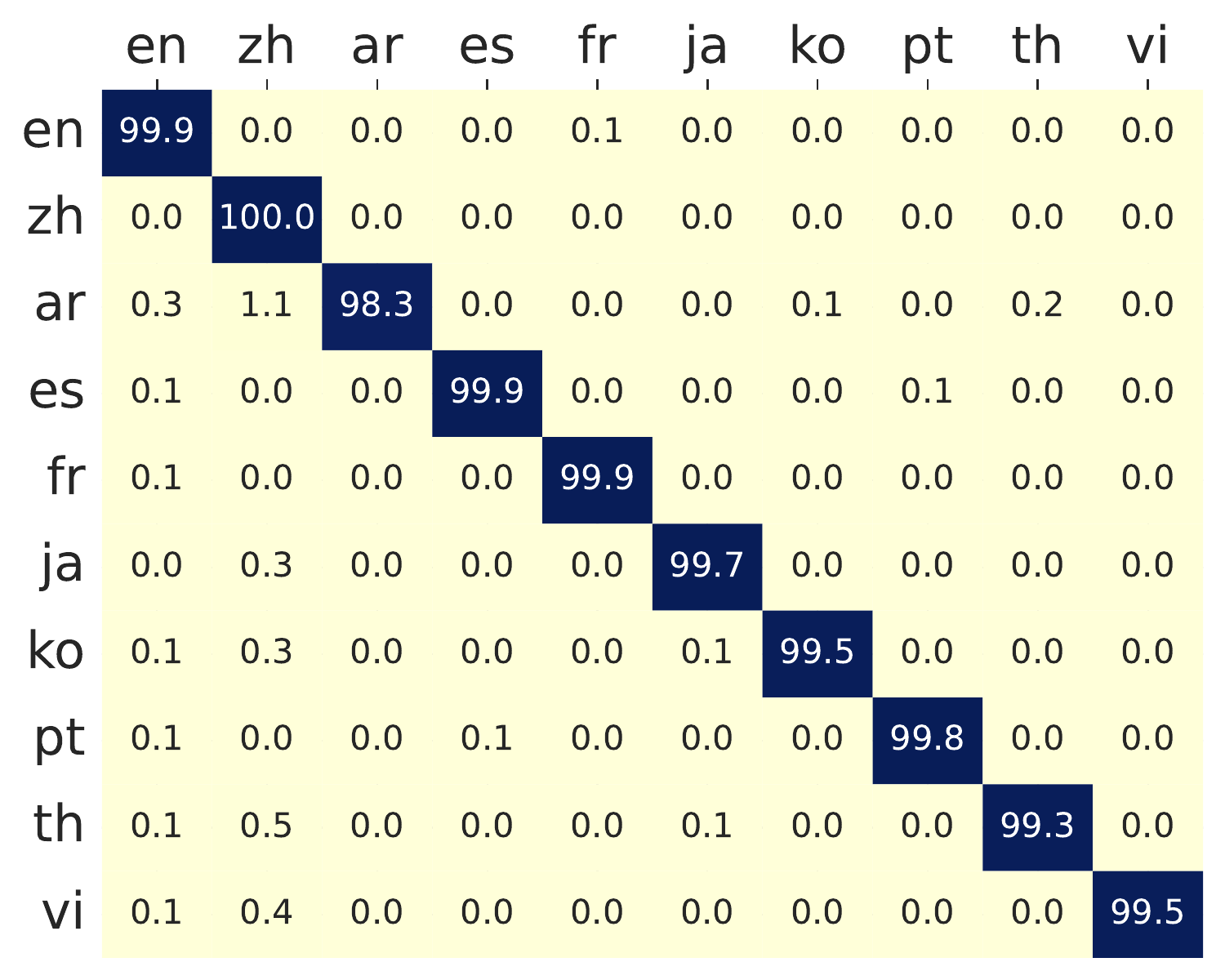}
\caption{Qwen2.5-32B-Instruct}
\end{subfigure}
\caption{The percentage to think in each language. The vertical is the source language and the horizontal is the target language. For Qwen2.5-32B-Instruct, as its response doesn't contain <think>, we use the whole response language instead.}
\label{fig:native_think}
\end{figure}

\begin{figure}[h]
\centering
\begin{subfigure}{0.49\linewidth}
\includegraphics[width=\linewidth]{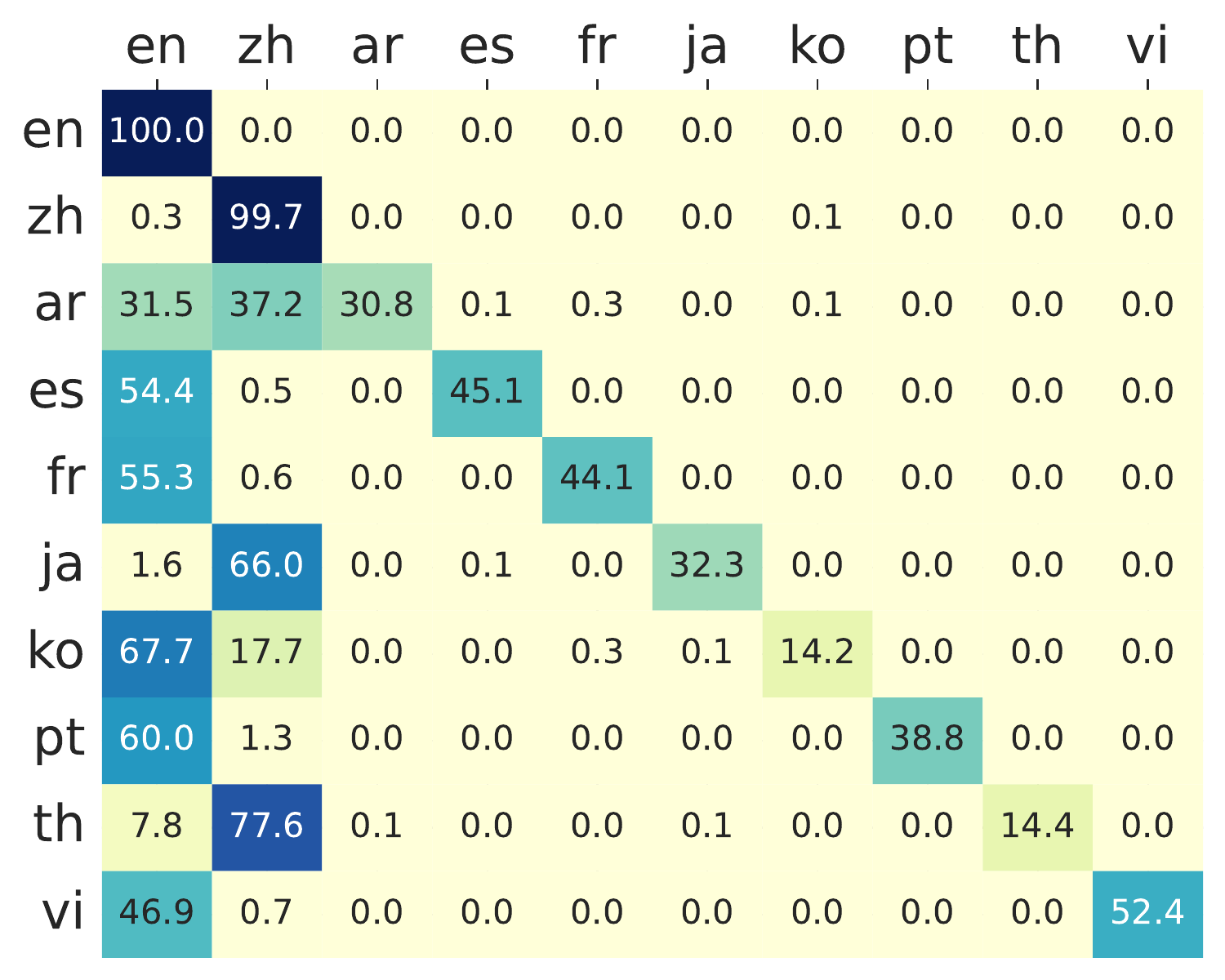}
\caption{Distill-Qwen-7B}
\end{subfigure}
\begin{subfigure}{0.49\linewidth}
\includegraphics[width=\linewidth]{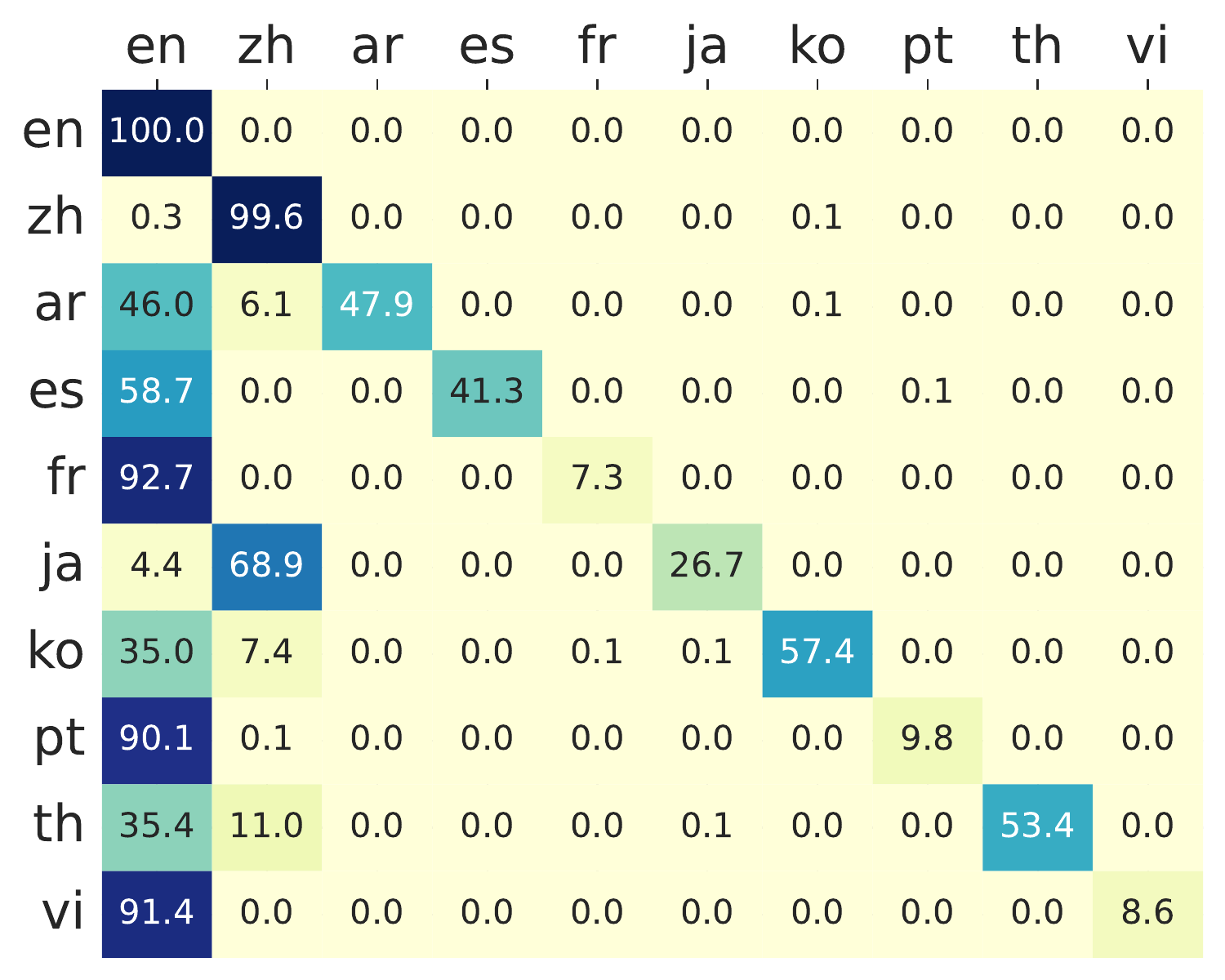}
\caption{Distill-Qwen-32B}
\end{subfigure}
\begin{subfigure}{0.49\linewidth}
\includegraphics[width=\linewidth]{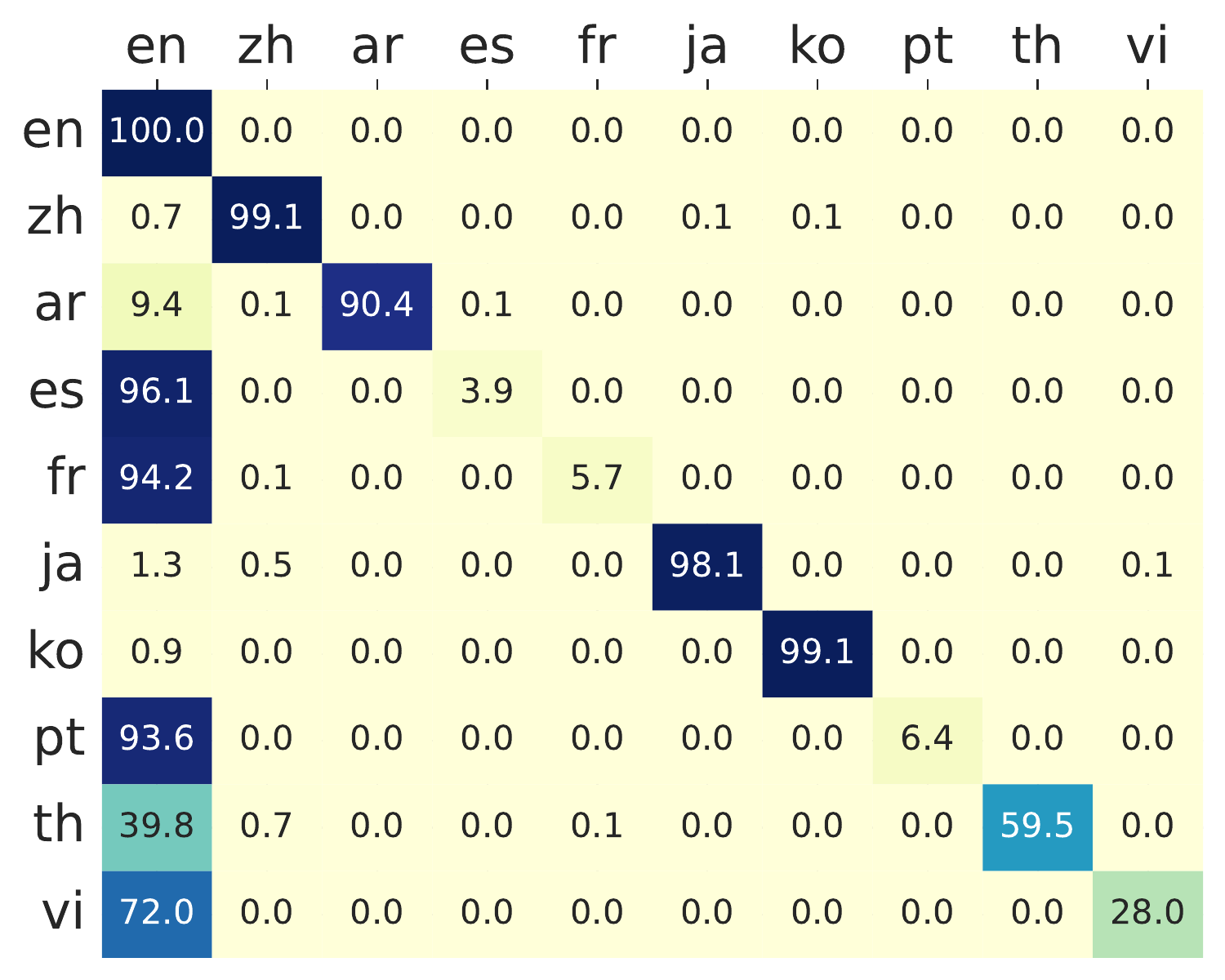}
\caption{QwQ-32B}
\end{subfigure}
\begin{subfigure}{0.49\linewidth}
\includegraphics[width=\linewidth]{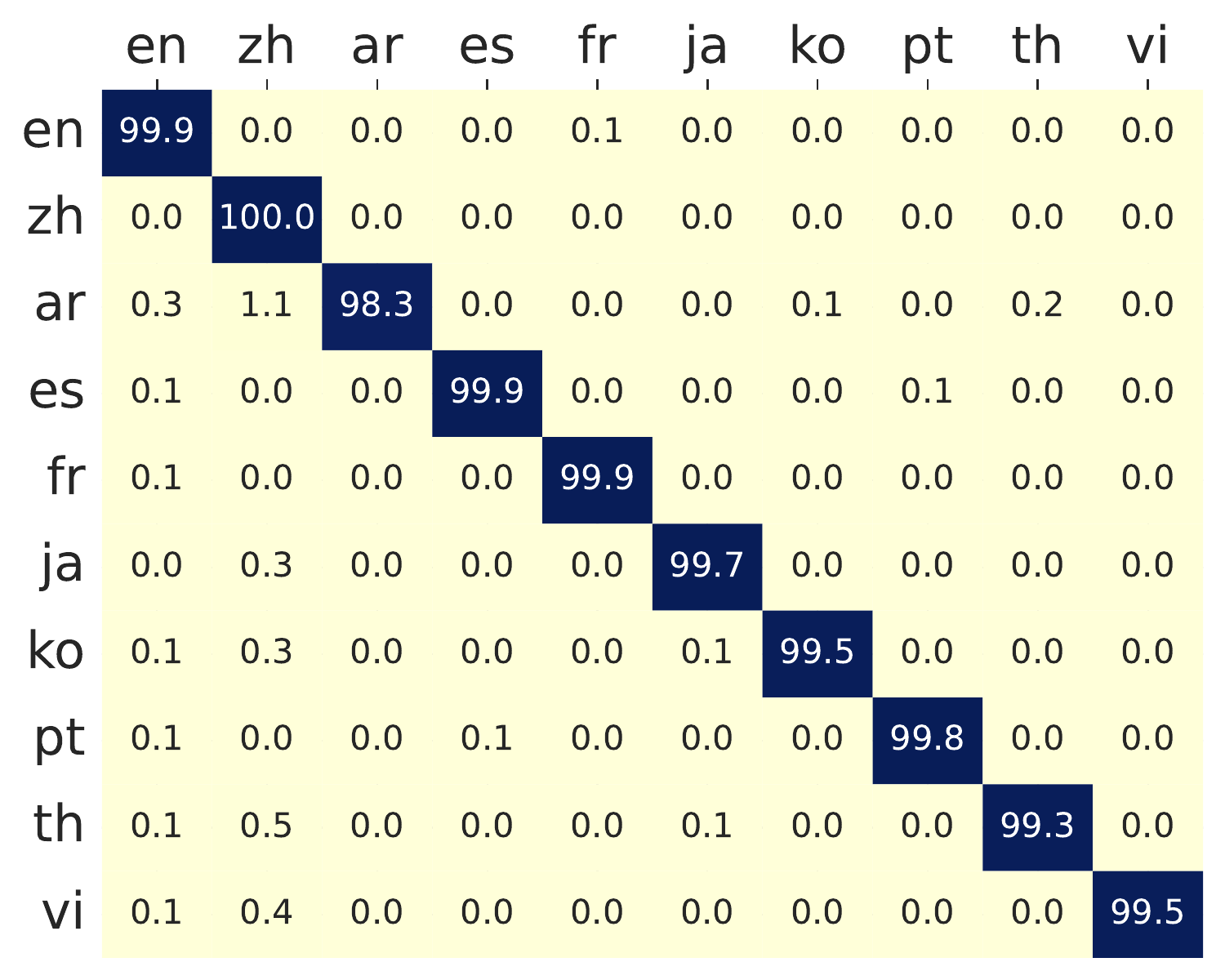}
\caption{Qwen2.5-32B-Instruct}
\end{subfigure}
\caption{The percentage to answer in each language.}
\label{fig:native_answer}
\end{figure}

Figure~\ref{fig:native_think} and~\ref{fig:native_answer} show some more specific results, the former shows how often a language is used in thinking steps and the latter shows the same statistics for the answer part. As we can see, for high-resource questions like English and Chinese, all models tend to answer in the native language. However, for low-resource languages like Arabic, DeepSeek-R1-Distill-Qwen-7B shows a tendency to answer in either English, Chinese, or Arabic in equal probability. This phenomenon varies between languages, as Thai tends to answer in Chinese while Vietnamese tends to English. QwQ-32B demonstrates relatively better language consistency, except in the cases of Spanish, French, and Portuguese, where it often defaults to English in both thinking and answering steps. We hypothesize that this behavior stems from post-training processes that heavily emphasize high-resource languages.


\paragraph{Off-target Thinking Increases Accuracy.}
To assess whether this off-target thinking (Figure~\ref{fig:native_think}) actually improves the accuracy of mathematical reasoning, we compare the results between thinking in the target language and thinking in off-target languages in Table~\ref{tab:target_vs_off}.

\begin{table}[htbp]
\centering
\scalebox{0.6}{
\begin{tabular}{lccccccccccc}
\toprule
Type & EN & ZH & AR & ES & FR & JA & KO & PT & TH & VI & AVG \\
\midrule
\multicolumn{12}{c}{DeepSeek-R1-Distill-Qwen-7B} \\
\cdashline{1-12}\noalign{\vskip 0.3ex}
Target & 57 & 45 & 0 & 60 & 33 & 0 & N/A & 50 & 0 & 25 & 30 \\
Off-target & N/A & 0 & 44 & 53 & 53 & 28 & 47 & 50 & 29 & 51 & 39 \\
\midrule
\multicolumn{12}{c}{DeepSeek-R1-Distill-Qwen-32B} \\
\cdashline{1-12}\noalign{\vskip 0.3ex}
Target & 72 & 48 & 42 & 48 & N/A & 0 & 30 & N/A & 22 & 0 & 33 \\
Off-target & N/A & 0 & 60 & 69 & 67 & 45 & 67 & 69 & 66 & 69 & 57 \\
\midrule
\multicolumn{12}{c}{QwQ-32B} \\
\cdashline{1-12}\noalign{\vskip 0.3ex}
Target & 77 & 65 & 67 & N/A & N/A & 56 & 59 & 33 & 65 & 71 & 62 \\
Off-target & N/A & N/A & 64 & 81 & 77 & 0 & 50 & 79 & 70 & 76 & 62 \\
\bottomrule
\end{tabular}
}
\caption{The accuracy between thinking in target language and off-target language. N/A means the model has no sample thinking in that language.}
\label{tab:target_vs_off}
\end{table}

From the results, we observe a consistent trend: off-target thinking often yields comparable or even superior accuracy compared to reasoning strictly in the target language. This phenomenon is even more observable in low-resource languages. For instance, in the DeepSeek-R1-Distill-Qwen-7B model, when tackling Arabic, the model completely fails when reasoning in the target language, yet achieves a substantial improvement (0.44) through off-target reasoning. A similar pattern appears in Thai. 

In high-resource languages, however, the benefit of off-target thinking is less pronounced. For example, for English, which often serves as the backbone language in pretraining, target-language reasoning yields the highest accuracies across all models (0.57, 0.72, and 0.77, respectively) and no off-target reasoning is recorded.



\begin{table*}[h]
\centering
\small
\scalebox{1.0}{
\begin{tabular}{lccccccccccc}
\toprule
Model & EN & ZH & AR & ES & FR & JA & KO & PT & TH & VI & AVG \\
\midrule
Distill-Qwen-7B & 63.90 & 58.53 & 56.50 & 62.81 & 61.58 & 50.90 & 59.90 & 62.72 & 48.97 & 58.53 & 58.44 \\
Distill-Qwen-7B-\atp & 62.64 & 56.61 & 51.67 & 64.97 & 62.59 & 40.62 & 58.66 & 62.64 & 49.31 & 51.61 & 56.13 \\
Distill-Qwen-7B-\dit & 62.63 & 55.88 & 56.38 & 52.61 & 49.68 & 24.53 & 52.58 & 50.30 & 40.22 & 38.53 & 48.34 \\
Distill-Qwen-7B-\qrt & 62.48 & 57.23 & 56.70 & 51.81 & 49.74 & 29.73 & 37.97 & 55.88 & 33.19 & 40.48 & 47.52 \\
\midrule
Distill-Qwen-32B & 73.94 & 61.69 & 65.02 & 71.96 & 70.88 & 60.29 & 59.23 & 72.68 & 63.31 & 71.12 & 67.01 \\
Distill-Qwen-32B-\atp & 73.35 & 59.06 & 68.56 & 69.38 & 72.51 & 66.18 & 69.91 & 72.77 & 66.18 & 70.08 & 68.80 \\
Distill-Qwen-32B-\dit & 72.29 & 61.99 & 58.71 & 65.79 & 63.48 & 37.85 & 49.42 & 62.72 & 51.78 & 47.04 & 57.11 \\
Distill-Qwen-32B-\qrt & 71.68 & 58.15 & 56.21 & 61.40 & 61.22 & 48.01 & 48.35 & 62.67 & 49.36 & 58.74 & 57.58 \\
\midrule
QwQ-32B & 79.43 & 74.72 & 71.10 & 80.27 & 79.04 & 64.38 & 68.56 & 78.65 & 73.43 & 77.28 & 74.69 \\
QwQ-32B-\atp & 78.95 & 72.85 & 79.27 & 78.52 & 78.80 & 66.07 & 68.57 & 78.31 & 78.23 & 73.78 & 75.34 \\
QwQ-32B-\dit & 78.34 & 75.47 & 69.68 & 76.00 & 74.12 & 68.60 & 67.56 & 75.54 & 68.39 & 71.64 & 72.53 \\
QwQ-32B-\qrt & 77.86 & 74.12 & 71.90 & 74.73 & 76.80 & 66.90 & 66.18 & 75.45 & 67.00 & 72.50 & 72.34 \\
\bottomrule
\end{tabular}}
\caption{Evaluation results of different evaluation strategies. \atp means prompting models to answer in the target language. \dit introduces multilingual discourse markers to induce models' thinking language. \qrt imitates models' behavior to repeat questions before thinking about how to solve them.}
\label{tab:eval_strategy}
\end{table*}

\begin{table}[htbp]
\centering
\small
\scalebox{0.95}{
\begin{tabular}{lcc}
\toprule
Model & Thinking LCR & Answering LCR \\
\midrule
Distill-Qwen-7B & 45.31 & 47.11 \\
Distill-Qwen-7B-\atp & 56.54 & 56.48 \\
Distill-Qwen-7B-\dit & 75.61 & 74.10 \\
Distill-Qwen-7B-\qrt & 81.41 & 78.04 \\
\midrule
Distill-Qwen-32B & 40.12 & 45.13 \\
Distill-Qwen-32B-\atp & 29.38 & 74.55 \\
Distill-Qwen-32B-\dit & 97.02 & 96.57 \\
Distill-Qwen-32B-\qrt & 97.73 & 97.62 \\
\midrule
QwQ-32B & 57.47 & 58.94 \\
QwQ-32B-\atp & 35.98 & 68.27 \\
QwQ-32B-\dit & 98.40 & 96.20 \\
QwQ-32B-\qrt & 99.88 & 97.86 \\
\bottomrule
\end{tabular}
}
\caption{LCR results for different elicitation strategies.}
\label{tab:eval_lcr_results}
\end{table}

\subsubsection{Explicit Language Elicitation}\label{sec:Explicit Language Elicitation}
In this section, we try different methods to elicit models to respond in the target language, aiming at mitigating the off-target problem.

\paragraph{Can Models Answer in the Target Language with Explicit Prompts?}
We begin by examining whether models can be guided to answer in the target language using explicit prompts. To leverage the models' internal English reasoning capabilities, we append a multilingual version of the instruction ``please think in English and answer in [target language]'' after the native language prompts. We refer to this strategy as the Answer-in-Target Prompt (\atp) as illustrated in Figure~\ref{fig:ans-in-target-prompt}. 

As shown in Table~\ref{tab:eval_strategy} and~\ref{tab:eval_lcr_results}, applying \atp slightly influences the accuracy results and answering LCR. For example, DeepSeek-R1-Distill-Qwen-32B and QwQ-32B have a 1\% accuracy increase, while DeepSeek-R1-Distill-Qwen-7B even experiences a performance decline. Considering LCR, we find that DeepSeek-R1-Distill-Qwen-7B has a comparable thinking and answering LCR, which indicates it may has already lost the ability to follow multilingual instructions, explaining why its accuracy decreases. Furthermore, the answering LCR of different models are greatly enhanced, indicating reasoning models are naturally possible to answer in the target language to some extent.


\paragraph{Can Thinking Intervention Mitigate Off-target Problem?}
Recent research on complex reasoning~\cite{wu-etal-2025-effectively,ma-etal-2025-reasoning} has proved that thinking intervention could provide a more fine-grained control over models' behavior. In this part, we collect several multilingual thinking patterns we observed in models' original thinking responses and see whether this could mitigate the off-target issue.

$\bullet$ \emph{Discourse-Initiated Thinking (\dit).}
When asked in English, the model tends to start their thinking with discourse markers like ``Alright'' or ``Okay'', we observe a similar pattern in multilingual scenarios as shown in Figure~\ref{fig:discourse_marks}. To leverage this behavior, we extract these markers from native prompt responses. When multiple candidates are available, one is randomly selected and appended after the <think> token. This approach encourages models to initiate their reasoning using discourse cues as entry points into the thinking process.

$\bullet$ \emph{Question-Restatement Thinking (\qrt).}
Another common pattern observed is that models often restate the question before engaging in actual reasoning. We replicate this behavior by explicitly inserting a restated version of the question at the beginning of the thinking process, as illustrated in Figure~\ref{fig:ques_repeat}. This intervention encourages the model to frame the problem before attempting to solve it.

As shown in Table~\ref{tab:eval_strategy}, \dit and \qrt lead to serious performance drops for the two distilled models, especially on languages except English and Chinese. Compared with them, QwQ shows a relatively better result, which might be attributed to the multilingual CoT training. However, answering LCR results in Table~\ref{tab:eval_lcr_results} have shown great increases, which means the models' off-target problem could be effectively mitigated with moderate thinking intervention, and these results may show more truthful multilingual ability for multilingual models.

\subsubsection{Training with English reasoning traces}\label{sec:training}
In this section, we manually create multilingual training datasets and further prove that training with English thinking could help increase performance while maintaining answering LCR.

\paragraph{Datasets Creation.}
To construct a moderately sized dataset for complex reasoning, we use the 3K-example subset of Light-R1~\cite{wen-etal-2025-light}. The questions are sourced from recent benchmarks such as Open-R1~\cite{openr1}, LIMO~\cite{ye-etal-2025-limoreasoning}, and S1~\cite{muennighoff-etal-2025-s1simpletesttimescaling}, while the answers are generated via knowledge distillation from DeepSeek-R1~\cite{guo-etal-2025-deepseek}. This dataset is filtered to retain 3,000 examples based on reasoning complexity, with all questions, thought processes, and answers presented in English.

Given the lack of multilingual datasets for complex reasoning, we translate this English dataset into 10 languages using \textit{GPT-4o-mini}\footnote{\url{https://openai.com/index/gpt-4o-mini-advancing-cost-efficient-intelligence/}}. To reduce translation inconsistencies in long-form content, we segment the reasoning process into paragraphs and translate each step-by-step.

\paragraph{Training Setups.}
Based on the constructed dataset, we propose three supervised fine-tuning strategies designed to enhance multilingual reasoning and maintain language consistency. Each method utilizes 3K different examples with the same prompts used in Section~\ref{sec:exp-setup}.

$\bullet$ \emph{\ensft.} We directly fine-tune on the original English dataset, where the question, thinking, and answering are all in English.

$\bullet$ \emph{\nativethink.} To ensure comparability with the English setup, we randomly divide the dataset indices into 10 parts (corresponding to the target languages), assigning each part to a specific language. This results in 300 examples per language. In this setting, the question, thinking, and answering are all in the respective native language.

$\bullet$ \emph{\enthink.} As discussed in Section~\ref{sec:off-target problem}, using English reasoning may enhance multilingual understanding while preserving answering LCR. Based on the \nativethink setup, we substitute the thinking component with its English version while retaining the native language for the question and answer parts.

We fine-tune Qwen2.5-32B-Instruct for 3 epochs on 8 H100 GPUs, employing a cosine learning rate schedule with a peak learning rate of $1 \times 10^{-5}$, a warm-up ratio of 0.1, and a total batch size of 8. And we report the results from the final checkpoint.

\paragraph{Results.}
The performance results are summarized in Table~\ref{tab:trained_results}, LCR scores are shown in Table~\ref{tab:train_lcr_results}, and the language usage for reasoning and answering is illustrated in Figure~\ref{fig:trained_results_think_ans_lagnuage}. As shown, all three training strategies yield substantial improvements over the base model (Qwen2.5-32B-Instruct). Specifically, the \ensft strategy improves performance to 62.38, demonstrating the effectiveness of supervised fine-tuning on English data. The \nativethink variant, which encourages the model to reason in the question’s native language, achieves a comparable average score of 61.46. Notably, the \enthink variant—where the model reasons in English regardless of the input language—achieves the highest average score of 66.72, outperforming all other configurations and approaching the performance of Distill-Qwen-32B (67.01).

In terms of language consistency, \ensft yields the lowest answering LCR, suggesting that English-only training contributes to off-target responses. While \nativethink increases answering LCR, it does not lead to notable performance gains. In contrast, \enthink maintains a high answering LCR (97.61) while significantly boosting performance. These findings provide strong evidence that reasoning in English while answering in the target language is an effective strategy for enhancing multilingual complex reasoning.

\begin{table*}[htbp]
\centering
\small
\scalebox{1.0}{%
\begin{tabular}{lccccccccccc}
\toprule
Model & EN & ZH & AR & ES & FR & JA & KO & PT & TH & VI & AVG \\
\midrule
Base & 38.43 & 29.38 & 27.03 & 31.13 & 29.48 & 25.94 & 26.44 & 31.17 & 27.76 & 27.37 & 29.41 \\
\ensft & 65.35 & 65.09 & 52.20 & 67.65 & 66.08 & 59.54 & 54.18 & 64.93 & 62.93 & 65.83 & 62.38 \\
\nativethink & 65.38 & 59.82 & 61.58 & 65.18 & 66.01 & 56.93 & 52.86 & 65.48 & 58.29 & 63.04 & 61.46 \\
\enthink & 66.00 & 66.31 & 65.82 & 68.44 & 66.37 & 65.90 & 67.87 & 66.11 & 66.29 & 68.10 & 66.72 \\
\bottomrule
\end{tabular}
}
\caption{Evaluation results of different training strategies on our benchmark based on Qwen2.5-32B-Instruct. AVG represents the average score across languages.}
\label{tab:trained_results}
\end{table*}

\begin{table}[htbp]
\centering
\small
\scalebox{1.0}{
\begin{tabular}{lcc}
\toprule
Model & Thinking LCR & Answering LCR\\
\midrule
\ensft & 57.69 & 59.20 \\
\nativethink & 99.85 & 99.76 \\
\enthink & 10.04 & 97.61 \\
\bottomrule
\end{tabular}}
\caption{Language Consistency Ratio (LCR) for different training setting for Qwen2.5-32B-Instruct.}
\label{tab:train_lcr_results}
\end{table}

\section{Related Work}\label{sec:related work}

\paragraph{Multilingual Reasoning.} Multilingual reasoning with large language models (LLMs) has received increasing attention, driven by the need for equitable performance across languages. For example, \citet{shi-etal-2022-language} build the first multilingual mathmatical reasoning benchmark, MGSM, based on GSM8k~\cite{cobbe-etal-2021-training} and provides several prompting strategies like \encot, which asks the model to predict the chain of thought in English.

Based on the benchmark, more and more techniques have been developed such as prompting and fine-tuning. For example, XLT~\cite{huang-etal-2023-not} prompts models to translate the question into English and solve it step-by-step, while \citet{liu-etal-2024-translation} leverage multilingual models like NLLB~\cite{costa-etal-2022-no} to improve translation quality. \citet{chen-etal-2023-breaking} further fine-tune models on multilingual data by training them to answer questions either in the same language or across different languages. With a further step, QAlign~\cite{zhu-etal-2024-question} explores the benefits of question alignment, where they explicitly train the model to translate reasoning questions into English.

Despite the advancement of previous research, most of them are based on MGSM, which appears too easy for temporary reasoning models (\eg Claude 3.5 Sonnet 91.6). To address the limitation, concurrent work like MCLM~\cite{son-etal-2025-linguistic} and PolyMath~\cite{wang-etal-2025-polymath} introduce new reasoning benchmarks across diverse languages. Compared with our work, we not only provide a multilingual complex reasoning benchmark with several subsets and diverse languages, but also focus on the problem of off-target, which is overlooked by previous work. Based on the analysis, we propose several strategies to balance the performance and the off-target phenomenon.

\paragraph{Complex Reasoning.} 
Solving complex reasoning tasks with LLMs is advancing rapidly, driven by methods that enhance test-time computation and learning dynamics. One line of work introduces step-level feedback through process reward models, which score intermediate reasoning steps~\cite{yuan-etal-2024-free, snell-etal-2024-scaling}. Another adopts planning-based techniques such as Monte Carlo tree search to explore and optimize reasoning paths~\cite{feng-etal-2023-alphazero, qi-etal-2024-mutual, guan-etal-2025-rstar}. DeepSeek-R1~\cite{guo-etal-2025-deepseek} shows that LLMs can develop strong reasoning skills through reinforcement learning with simple rule-based rewards, without intermediate supervision. Follow-up studies~\cite{OpenReasonerZero2025, openr1} extend this approach to open-source models. Despite these advances, most work remains focused on English benchmarks like AIME and MATH-500~\cite{lightman-etal-2023-let}, overlooking the multilingual aspect. This work addresses that gap by introducing \mmath, a benchmark for complex reasoning across diverse languages.

\section{Conclusion}
In this paper, we introduce \mmath, a new multilingual benchmark to evaluate models' complex reasoning ability. \mmath is an extension of the widely used benchmark including AIME, CNMO and MATH-500. It contains 374 examples in 10 typologically diverse languages. Then we present a comprehensive analysis of the multilingual complex reasoning abilities of large reasoning models. We find that temporary reasoning models still show a gap in low-resource language scenarios. Finally, we propose several strategies like prompting, thinking intervention, and training, revealing the possibility to utilize models' English reasoning ability to enhance their multilingual performance while maintaining language consistency.


\section*{Limitations}
In this paper, we introduce a new multilingual benchmark for complex reasoning and conduct empirical studies on its effectiveness. Nonetheless, several challenges remain as limitations of our work.
First, although we explore training-free approaches, balancing accuracy and language consistency remains a significant challenge. Further investigation is needed to develop strategies that optimize both aspects. 
Secondly, synthesizing multilingual reasoning data remains a challenging problem. Our translation-based approach represents a preliminary attempt in this direction. Future work can explore additional methods for generating native multilingual reasoning data, encompassing both multilingual reasoning processes and corresponding answers.
Furthermore, our work focuses on mathematical reasoning. However, it leaves various tasks (\eg coding and STEM) within the context of multilingualism for large reasoning models unexplored.



\bibliography{ref}
\clearpage
\appendix

\section{Prompts in Benchmark Creation}
To construct a reliable multilingual benchmark, we designed prompts that guide models to translate mathematical questions accurately and evaluate the quality of these translations. This section presents the prompts used for both translation and translation revision, as shown in Table~\ref{tab:translation prompt} and Table~\ref{tab:judge prompt}.

\begin{table}[htbp]
\small
\centering
\begin{tabular}{p{0.95\columnwidth}}
    \toprule
    \begin{CJK*}{UTF8}{gbsn} 
I have the following mathematical question written in English, which contains LaTeX formatting. Please translate it into \{target\_lang\} while preserving the LaTeX format and mathematical notation. Ensure that the translation remains accurate and the mathematical expressions do not change. Do not add anything else. 

\#\#\# Example input:

How many positive whole-number divisors does 196 have?

\#\#\# Example output:

196有多少个正整数因子？

\#\#\# Input: 

\{text\}

\#\#\# Output:
    \end{CJK*}   \\
    \bottomrule
\end{tabular}
\caption{The prompt to translate English questions into other languages.}
\label{tab:translation prompt}
\end{table}

\begin{table}[htbp]
\small
\centering
\begin{tabular}{p{0.95\columnwidth}}
    \toprule
    \begin{CJK*}{UTF8}{gbsn} 
You are given two versions of a mathematical question: one in English and one in  \{target\_lang\} language (translated version). Your task is to evaluate whether the translated version is an accurate representation of the original English version. If the translation is correct, confirm that the translation is accurate and return 'Correct'. If the translation is incorrect or the language is wrong or there is unnessary parts, return 'Incorrect' and provide a corrected translation between <trans> and </trans>.

Analyse step by step.

\#\#\# Input:

English question: \{text\_en\}

Translated \{target\_lang\} question: \{text\}

\#\#\# Output:
    \end{CJK*}   \\
    \bottomrule
\end{tabular}
\caption{The prompt to judge translation results and give better feedback.}
\label{tab:judge prompt}
\end{table}

\section{Human Evaluation Details}
To ensure the correctness, fluency, and cultural appropriateness of translations in our multilingual benchmark, we conducted a comprehensive human evaluation. We recruited qualified validators with strong linguistic backgrounds to review the model-generated translations. Specifically, we engaged native speakers for Chinese, and university students for Vietnamese and Portuguese. For other languages, we selected individuals with corresponding language certifications. The validator details are summarized in Table~\ref{tab:human evaluations}.

Validators were instructed to evaluate the translations based on the accuracy of mathematical meaning, correctness of LaTeX formatting, and naturalness of language usage. Each validator was compensated with 1 \$ per example, and the whole evaluation process lasted 8 hours. Finally, we manually reviewed and carefully consolidated their assessments to ensure high-quality results.

\begin{table}[h]
\small
\centering
\begin{tabular}{|l|l|}
\hline
\textbf{Language} & \textbf{Language Certification or Identity} \\
\hline
Chinese & Native Speaker \\
Japanese & JLPT N1, TEM8 \\
French & TCF C1, TEM8 \\
Arabic & TEM8 \\
Spanish & MCER B2, TEM8 \\
Korean & TOPIK II \\
Vietnamese & Vietnam National University Student \\
Portuguese & Universidade NOVA de Lisboa Student \\
Thai & CUTFL Chula Superior \\
\hline
\end{tabular}
\caption{Languages and corresponding certifications (or identity) of translation validators.}
\label{tab:human evaluations}
\end{table}

\section{Benchmark Creation Process}
\begin{figure}[h]
\centering
\includegraphics[scale=1.0]{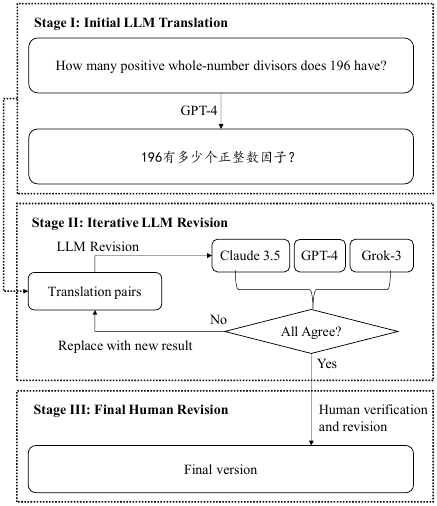}
\caption{The demonstration of our benchmark construction process.}
\label{fig:demo}
\end{figure}

\section{Case study}

\begin{table*}[htbp]
\small
\centering
\begin{tabular}{p{0.95\columnwidth}}
    \toprule
    \textbf{Question} \\
    \begin{CJK*}{UTF8}{gbsn} \textcolor{blue}{毎朝、アヤは$9$キロメートルの散歩に出かけてからコーヒーショップに立ち寄ります。彼女が毎時$s$キロメートルの一定の速度で歩くと、コーヒーショップで過ごす$t$分を含めて散歩には4時間かかります。彼女が毎時$s+2$キロメートルで歩くと、散歩には2時間24分かかり、その中にはコーヒーショップで過ごす$t$分が含まれます。アヤが毎時$s+\frac{1}{2}$キロメートルで歩くと仮定すると、コーヒーショップで過ごす$t$分を含めた散歩にかかる時間を分で求めなさい。}
    \end{CJK*} \\
    \midrule
    \textbf{Response} \\
    \begin{CJK*}{UTF8}{gbsn} <think>\textcolor{red}{好，我现在来解决这个问题。首先，我要理清楚题目中的各个条件和变量，然后逐步分析，找到正确的解题方法。
题目说，阿雅每天早上散步9公里，然后去咖啡店。她的速度是每小时s公里，包括在咖啡店停留的t分钟，整个过程需要4个小时。当她以s+2公里每小时的速度行走时，整个过程用了2小时24分钟，同样包括t分钟在咖啡店。现在，假设她以s+1/2公里每小时的速度行走，求包括t分钟在内的总时间，用分钟表示。\ldots现在，第三种情况：s+0.5=3公里/小时，步行时间=9/3=3小时=180分钟，加上t=24分钟，总时间=204分钟。
所以，答案应该是204分钟。}
</think>

\textcolor{blue}{阿ヤが毎時$s+\frac{1}{2}$キロメートルで歩く場合、コーヒーショップでの滞在を含めた散歩にかかる時間は204分です。}
    \end{CJK*} \\
    \bottomrule                                            
\end{tabular}
\caption{An example of cross-lingual thinking and answering. The blue text is Japanese, and the red text is Chinese.}
\label{tab:cross-lingual case}
\end{table*}




\section{Other Results}

\begin{table*}[htbp]
\centering
\small
\scalebox{1.0}{%
\begin{tabular}{lccccccccccc}
\toprule
Model & EN & ZH & AR & ES & FR & JA & KO & PT & TH & VI & AVG \\
\midrule
\multicolumn{12}{c}{AIME2024} \\
\cdashline{1-12}\noalign{\vskip 0.3ex}
Qwen2.5-32B-Instruct & 16.67 & 15.83 & 10.00 & 12.50 & 9.17 & 8.33 & 7.50 & 12.50 & 11.67 & 10.83 & 11.50 \\
Gemma3-27B-IT & 32.50 & 24.17 & 21.67 & 27.50 & 27.50 & 20.83 & 23.33 & 24.17 & 18.33 & 15.83 & 23.58 \\
DeepSeek-R1-Distill-Qwen-1.5B & 28.33 & 20.83 & 13.33 & 25.00 & 25.00 & 12.50 & 19.17 & 24.17 & 15.00 & 25.83 & 20.92 \\
DeepSeek-R1-Distill-Qwen-7B & 56.67 & 44.17 & 43.33 & 53.33 & 50.83 & 27.50 & 46.67 & 50.00 & 28.33 & 49.17 & 45.00 \\
DeepSeek-R1-Distill-Llama-8B & 43.33 & 25.83 & 23.33 & 45.00 & 39.17 & 18.33 & 23.33 & 40.83 & 16.67 & 40.00 & 31.58 \\
DeepSeek-R1-Distill-Qwen-14B & 69.17 & 45.83 & 64.17 & 62.50 & 64.17 & 44.17 & 54.17 & 60.00 & 59.17 & 61.67 & 58.50 \\
DeepSeek-R1-Distill-Qwen-32B & 71.67 & 47.50 & 55.00 & 65.00 & 66.67 & 44.17 & 48.33 & 69.17 & 50.83 & 68.33 & 58.67 \\
QwQ-32B & 76.67 & 65.00 & 66.67 & 80.83 & 76.67 & 55.83 & 59.17 & 77.50 & 68.33 & 75.00 & 70.17 \\
Deepseek-R1 & 76.67 & 70.00 & 73.33 & 79.17 & 76.67 & 67.50 & 65.00 & 78.33 & 70.00 & 78.33 & 73.50 \\
o3-mini & 80.83 & 75.83 & 75.83 & 76.67 & 77.50 & 76.67 & 70.83 & 80.00 & 71.67 & 78.33 & 76.42 \\
\midrule
\multicolumn{12}{c}{AIME2025} \\
\cdashline{1-12}\noalign{\vskip 0.3ex}
Qwen2.5-32B-Instruct & 15.00 & 8.33 & 6.67 & 11.67 & 11.67 & 10.00 & 8.33 & 15.00 & 6.67 & 13.33 & 10.67 \\
Gemma3-27B-IT & 25.00 & 31.67 & 30.00 & 30.00 & 26.67 & 26.67 & 30.00 & 21.67 & 31.67 & 23.33 & 27.67 \\
DeepSeek-R1-Distill-Qwen-1.5B & 28.33 & 16.67 & 18.33 & 25.00 & 20.00 & 18.33 & 16.67 & 13.33 & 8.33 & 23.33 & 18.83 \\
DeepSeek-R1-Distill-Qwen-7B & 38.33 & 40.00 & 36.67 & 41.67 & 40.00 & 38.33 & 43.33 & 46.67 & 25.00 & 40.00 & 39.00 \\
DeepSeek-R1-Distill-Llama-8B & 28.33 & 33.33 & 13.33 & 23.33 & 35.00 & 18.33 & 21.67 & 35.00 & 8.33 & 26.67 & 24.33 \\
DeepSeek-R1-Distill-Qwen-14B & 50.00 & 28.33 & 38.33 & 51.67 & 45.00 & 33.33 & 35.00 & 45.00 & 33.33 & 46.67 & 40.67 \\
DeepSeek-R1-Distill-Qwen-32B & 55.00 & 45.00 & 45.00 & 56.67 & 51.67 & 50.00 & 40.00 & 55.00 & 40.00 & 53.33 & 49.17 \\
QwQ-32B & 66.67 & 63.33 & 50.00 & 66.67 & 66.67 & 40.00 & 51.67 & 61.67 & 56.67 & 61.67 & 58.50 \\
Deepseek-R1 & 65.00 & 55.00 & 55.00 & 65.00 & 56.67 & 56.67 & 55.00 & 61.67 & 61.67 & 60.00 & 59.17 \\
o3-mini & 71.67 & 71.67 & 75.00 & 70.00 & 63.33 & 75.00 & 66.67 & 73.33 & 66.67 & 71.67 & 70.50 \\
\midrule
\multicolumn{12}{c}{CNMO} \\
\cdashline{1-12}\noalign{\vskip 0.3ex}
Qwen2.5-32B-Instruct & 36.11 & 12.50 & 13.89 & 16.67 & 13.89 & 9.72 & 12.50 & 13.89 & 16.67 & 6.94 & 15.28 \\
Gemma3-27B-IT & 51.39 & 38.89 & 34.72 & 34.72 & 41.67 & 36.11 & 33.33 & 36.11 & 33.33 & 38.89 & 37.92 \\

DeepSeek-R1-Distill-Qwen-1.5B & 37.50 & 30.56 & 43.06 & 37.50 & 45.83 & 40.28 & 34.72 & 33.33 & 29.17 & 29.17 & 36.11 \\
DeepSeek-R1-Distill-Qwen-7B & 65.28 & 58.33 & 61.11 & 63.89 & 65.28 & 54.17 & 65.28 & 62.50 & 56.94 & 61.11 & 61.39 \\
DeepSeek-R1-Distill-Llama-8B & 61.11 & 38.89 & 34.72 & 51.39 & 54.17 & 44.44 & 36.11 & 56.94 & 31.94 & 44.44 & 45.42 \\
DeepSeek-R1-Distill-Qwen-14B & 72.22 & 52.78 & 63.89 & 69.44 & 61.11 & 55.56 & 66.67 & 68.06 & 66.67 & 68.06 & 64.44 \\
DeepSeek-R1-Distill-Qwen-32B & 72.22 & 61.11 & 66.67 & 70.83 & 69.44 & 54.17 & 56.94 & 70.83 & 72.22 & 68.06 & 66.25 \\
QwQ-32B & 76.39 & 73.61 & 72.22 & 76.39 & 75.00 & 68.06 & 69.44 & 77.78 & 75.00 & 76.39 & 74.03 \\
Deepseek-R1 & 76.39 & 73.61 & 66.67 & 76.39 & 73.61 & 70.83 & 69.44 & 79.17 & 75.00 & 75.00 & 73.61 \\
o3-mini & 79.17 & 79.17 & 80.56 & 73.61 & 79.17 & 65.28 & 61.11 & 83.33 & 76.39 & 79.17 & 75.69 \\
\midrule
\multicolumn{12}{c}{MATH500} \\
\cdashline{1-12}\noalign{\vskip 0.3ex}
Qwen2.5-32B-Instruct & 85.93 & 80.87 & 77.57 & 83.68 & 83.20 & 75.72 & 77.41 & 83.28 & 76.05 & 78.38 & 80.21 \\
Gemma3-27B-IT & 93.33 & 90.84 & 88.91 & 92.12 & 91.96 & 88.42 & 88.10 & 91.48 & 88.26 & 90.19 & 90.36 \\

DeepSeek-R1-Distill-Qwen-1.5B & 87.46 & 82.32 & 63.26 & 74.12 & 77.49 & 69.21 & 67.04 & 71.14 & 59.73 & 69.21 & 72.10 \\
DeepSeek-R1-Distill-Qwen-7B & 95.34 & 91.64 & 84.89 & 92.36 & 90.19 & 83.60 & 84.32 & 91.72 & 85.61 & 83.84 & 88.35 \\
DeepSeek-R1-Distill-Llama-8B & 92.44 & 84.73 & 63.34 & 90.03 & 89.71 & 75.72 & 63.75 & 88.91 & 63.83 & 81.67 & 79.41 \\
DeepSeek-R1-Distill-Qwen-14B & 96.14 & 93.41 & 92.44 & 93.09 & 90.84 & 89.55 & 88.34 & 92.77 & 90.27 & 91.00 & 91.78 \\
DeepSeek-R1-Distill-Qwen-32B & 96.86 & 93.17 & 93.41 & 95.34 & 95.74 & 92.85 & 91.64 & 95.74 & 90.19 & 94.77 & 93.97 \\
QwQ-32B & 97.99 & 96.95 & 95.50 & 97.19 & 97.83 & 93.65 & 93.97 & 97.67 & 93.73 & 96.06 & 96.05 \\
Deepseek-R1 & 97.19 & 97.51 & 95.34 & 97.59 & 97.27 & 95.74 & 96.06 & 97.19 & 95.50 & 96.38 & 96.58 \\
o3-mini & 97.03 & 97.11 & 96.86 & 97.83 & 98.07 & 95.90 & 96.38 & 98.31 & 95.90 & 96.30 & 96.97 \\
\midrule
\multicolumn{12}{c}{MMATH} \\
\cdashline{1-12}\noalign{\vskip 0.3ex}
Qwen2.5-32B-Instruct & 38.43 & 29.38 & 27.03 & 31.13 & 29.48 & 25.94 & 26.44 & 31.17 & 27.76 & 27.37 & 29.41 \\
Gemma3-27B-IT & 50.55 & 46.39 & 43.82 & 46.09 & 46.95 & 43.01 & 43.69 & 43.36 & 42.90 & 42.06 & 44.88 \\
DeepSeek-R1-Distill-Qwen-1.5B & 45.41 & 37.59 & 34.50 & 40.40 & 42.08 & 35.08 & 34.40 & 35.49 & 28.06 & 36.89 & 36.99 \\
DeepSeek-R1-Distill-Qwen-7B & 63.90 & 58.53 & 56.50 & 62.81 & 61.58 & 50.90 & 59.90 & 62.72 & 48.97 & 58.53 & 58.44 \\
DeepSeek-R1-Distill-Llama-8B & 56.31 & 45.70 & 33.68 & 52.44 & 54.51 & 39.21 & 36.21 & 55.42 & 30.19 & 48.20 & 45.19 \\
DeepSeek-R1-Distill-Qwen-14B & 71.88 & 55.09 & 64.71 & 69.17 & 65.28 & 55.65 & 61.04 & 66.46 & 62.36 & 66.85 & 63.85 \\
DeepSeek-R1-Distill-Qwen-32B & 73.94 & 61.69 & 65.02 & 71.96 & 70.88 & 60.29 & 59.23 & 72.68 & 63.31 & 71.12 & 67.01 \\
QwQ-32B & 79.43 & 74.72 & 71.10 & 80.27 & 79.04 & 64.38 & 68.56 & 78.65 & 73.43 & 77.28 & 74.69 \\
Deepseek-R1 & 78.81& 74.03& 72.59& 79.54& 76.05& 72.69& 71.38& 79.09& 75.54& 77.43& 75.72 \\
o3-mini & 82.18& 80.95& 82.06& 79.53& 79.52& 78.21& 73.75& 83.74& 77.66& 81.37& 79.90 \\
\bottomrule
\end{tabular}
}
\caption{Evaluation results of different models on various subsets. Scores of \mmath are calculated with macro-average metric.}
\label{tab:main_results_all}
\end{table*}

\begin{figure*}[htbp]
\centering
\includegraphics[scale=0.6]{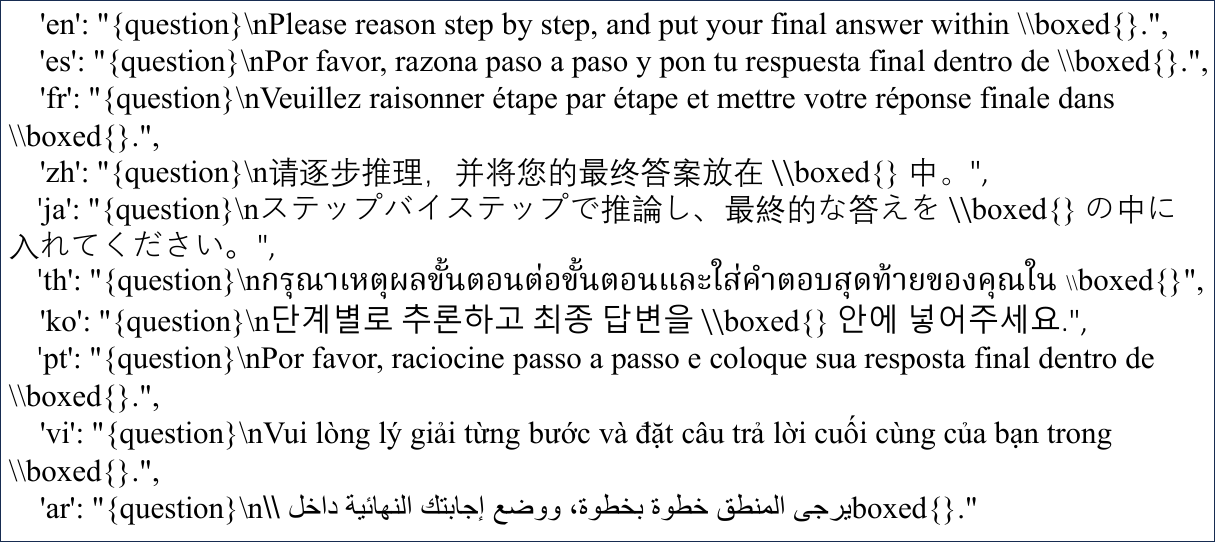}
\caption{Multilingual native language prompts for different languages.}
\label{fig:native-prompt}
\end{figure*}

\begin{figure*}[htbp]
\centering
\includegraphics[scale=0.7]{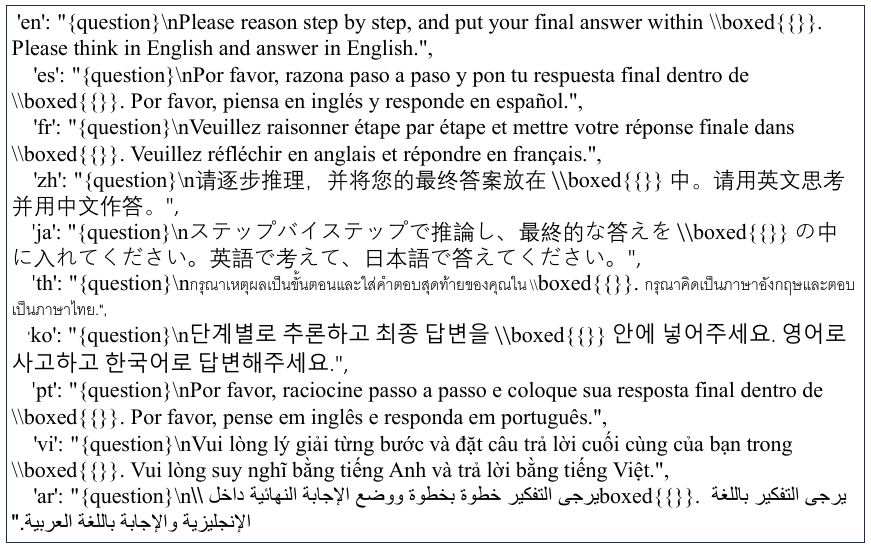}
\caption{Our \atp prompts, used to ask LLMs to explicitly answer in the target language.}
\label{fig:ans-in-target-prompt}
\end{figure*}

\begin{figure*}[htbp]
\centering
\includegraphics[scale=0.45]{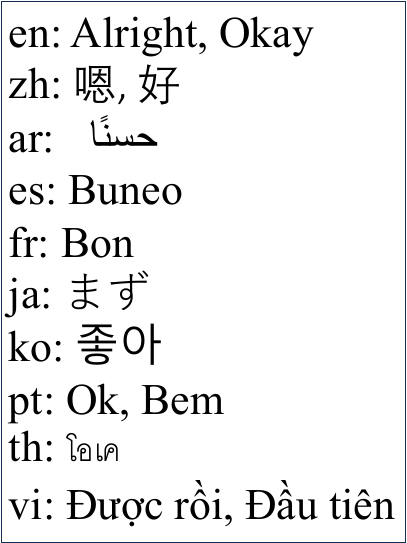}
\caption{Multilingual discourse marks used in our \dit thinking intervention method. These are collected from our observations about LLMs' native responses.}
\label{fig:discourse_marks}
\end{figure*}

\begin{figure*}[htbp]
\centering
\begin{subfigure}{0.49\linewidth}
\includegraphics[width=\linewidth]{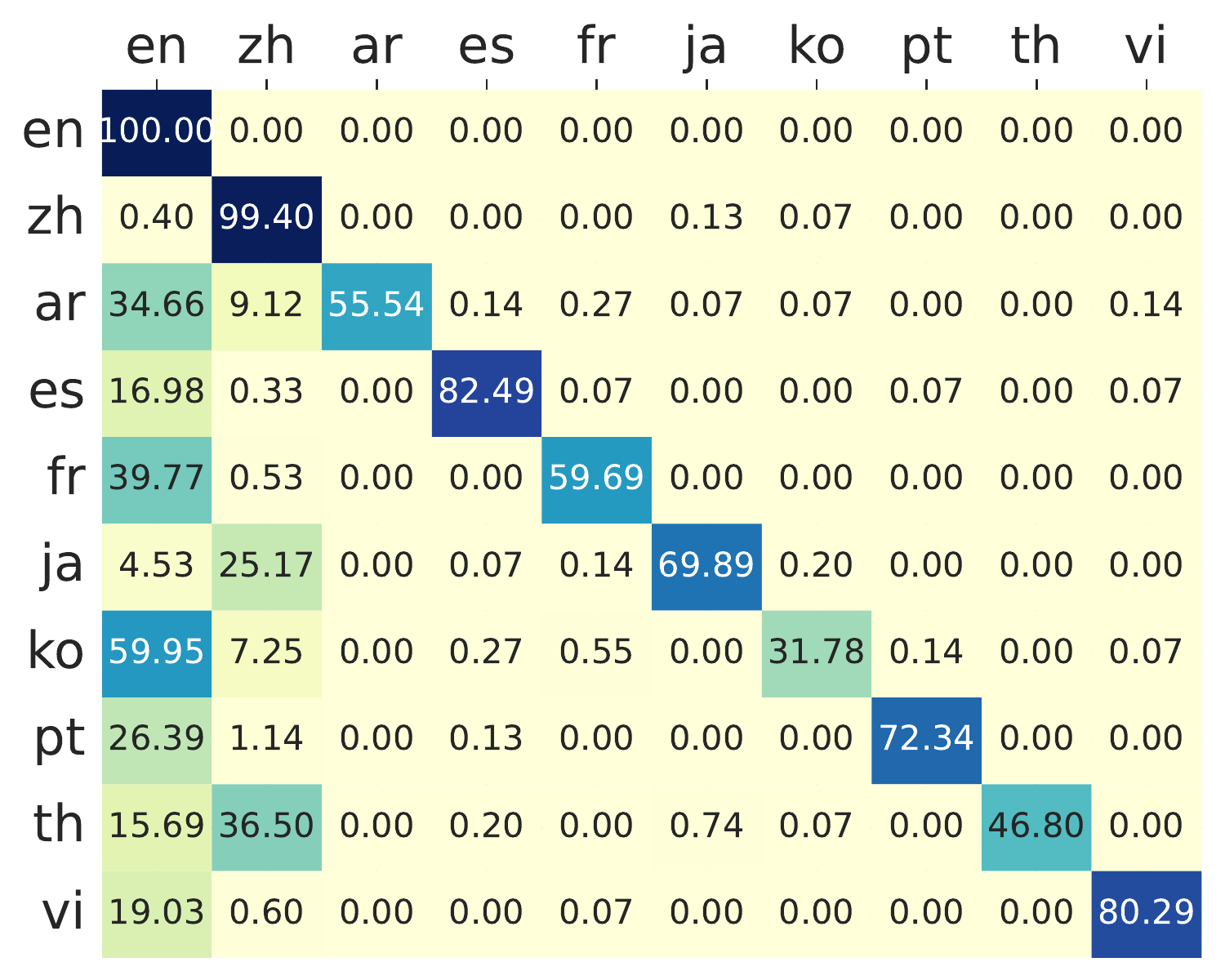}
\caption{DeepSeek-R1-Distill-Qwen-7B}
\end{subfigure}
\begin{subfigure}{0.49\linewidth}
\includegraphics[width=\linewidth]{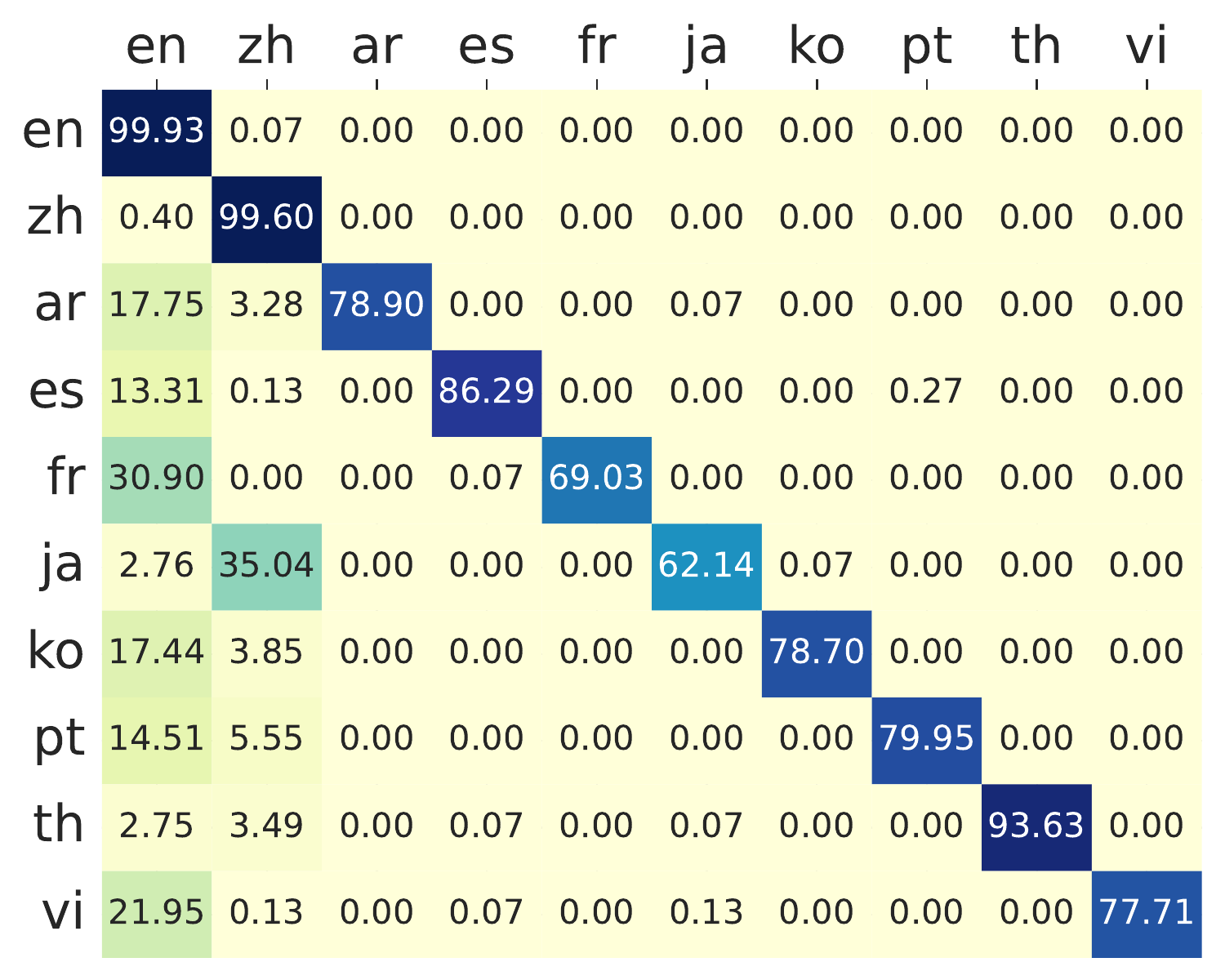}
\caption{DeepSeek-R1-Distill-Qwen-32B}
\end{subfigure}
\begin{subfigure}{0.49\linewidth}
\includegraphics[width=\linewidth]{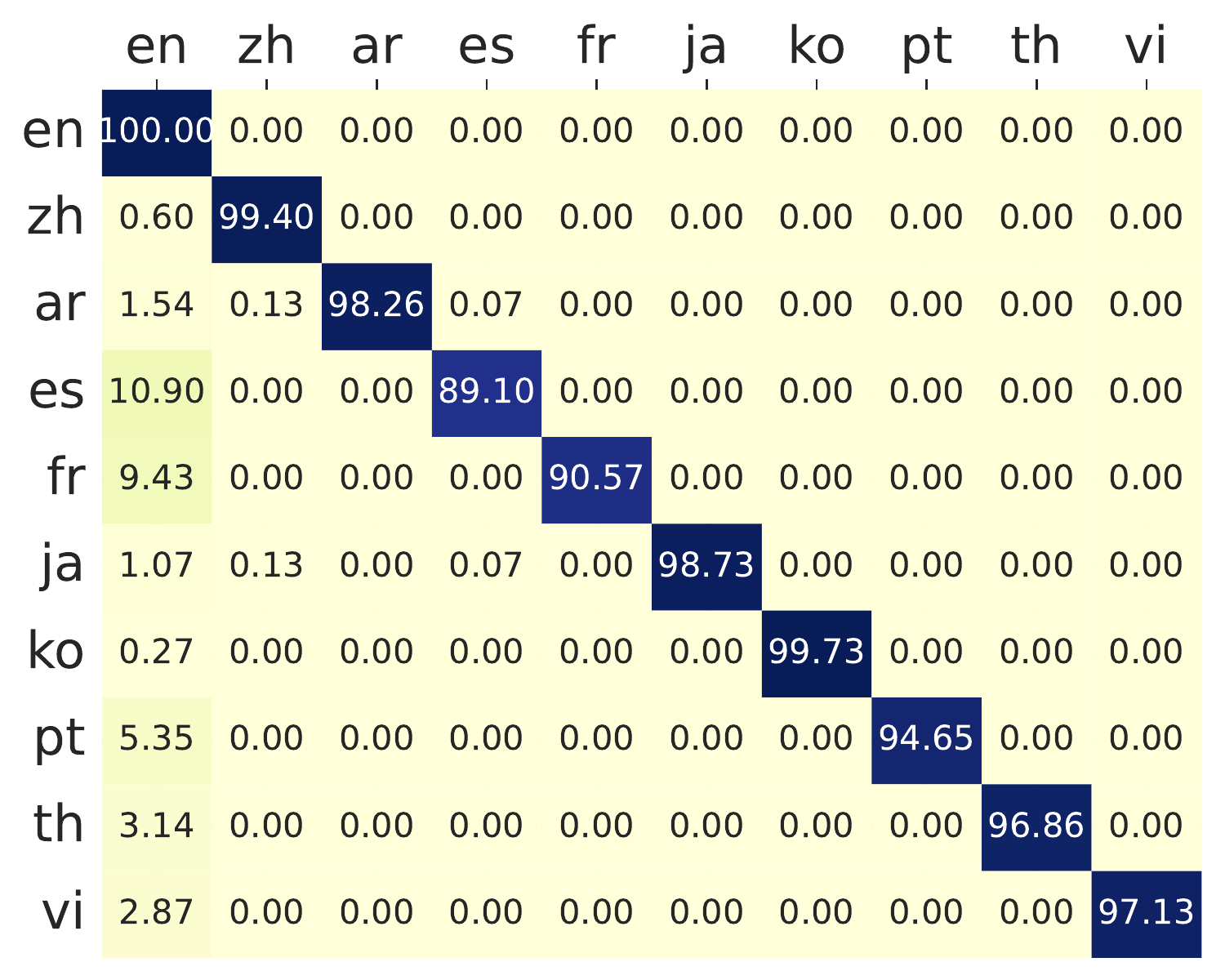}
\caption{QwQ-32B}
\end{subfigure}
\begin{subfigure}{0.49\linewidth}
\includegraphics[width=\linewidth]{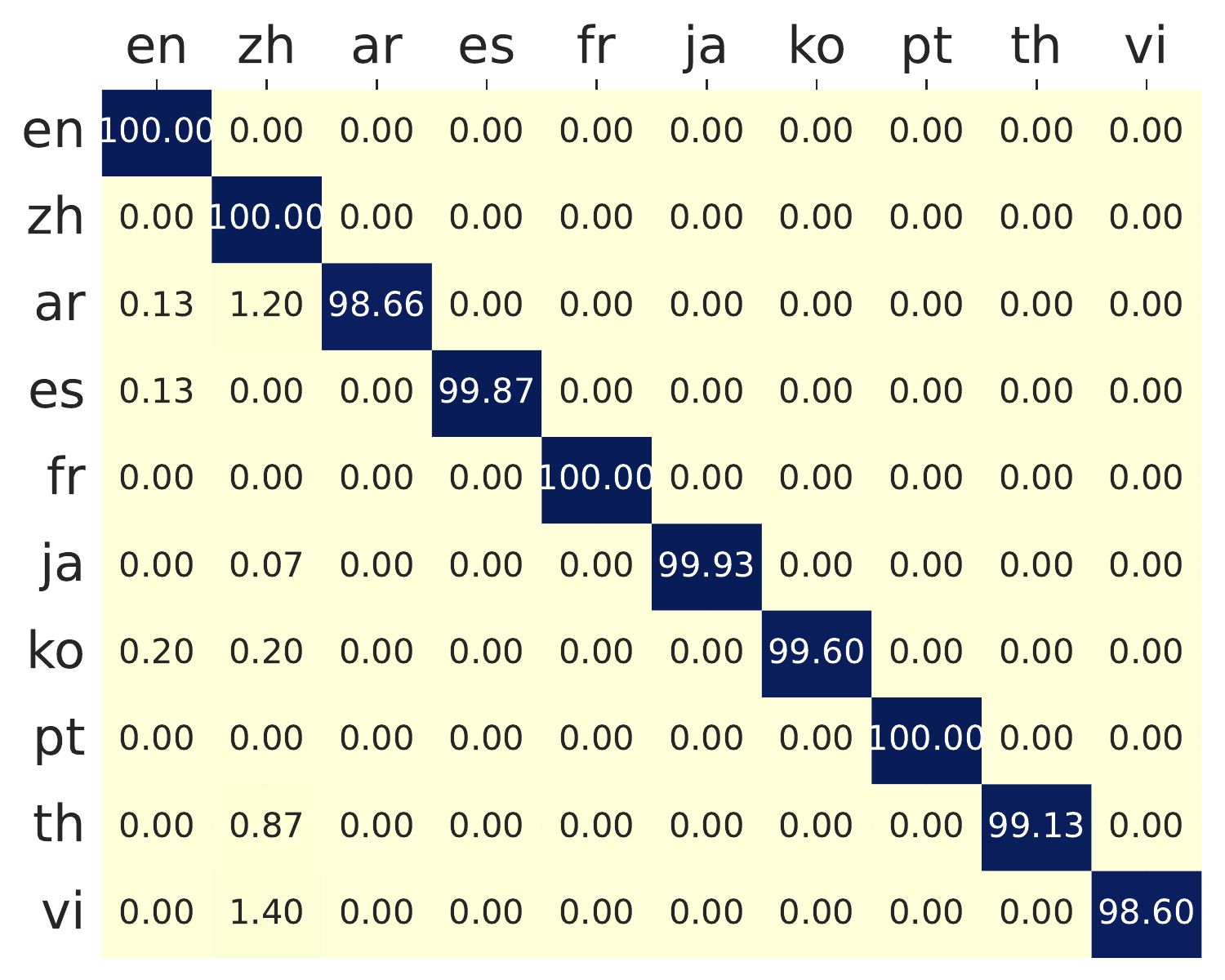}
\caption{Qwen2.5-32B-Instruct}
\end{subfigure}
\caption{The percentage to answer in each language when prompted with \atp (Figure \ref{fig:ans-in-target-prompt}).}
\label{fig:ans_in_tar_answer}
\end{figure*}

\begin{figure*}[htbp]
\centering
\includegraphics[scale=0.7]{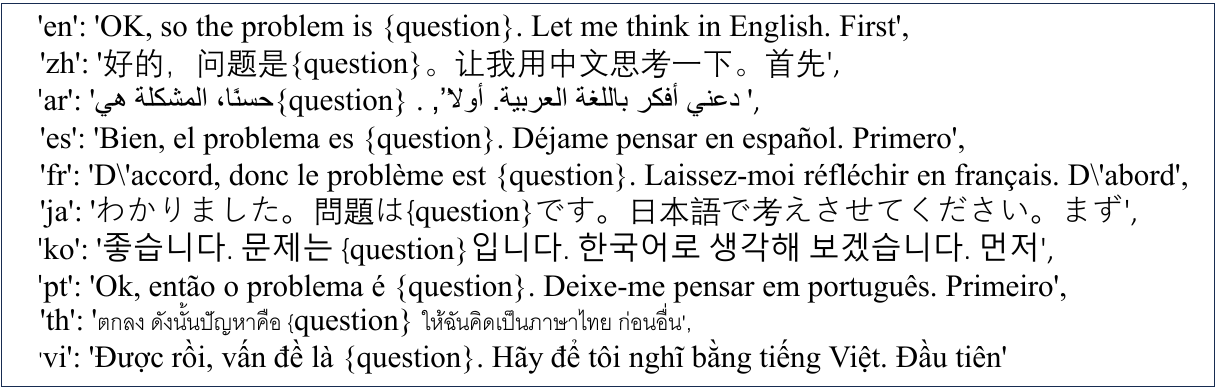}
\caption{Our \qrt thinking intervention, which imitates LLMs' behavior about repeating questions before actually thinking about how to solve it.}
\label{fig:ques_repeat}
\end{figure*}

\begin{table*}[htbp]
\centering
\small
\scalebox{1.0}{%
\begin{tabular}{lccccccccccc}
\toprule
Model & EN & ZH & AR & ES & FR & JA & KO & PT & TH & VI & AVG \\
\midrule
\multicolumn{12}{c}{AIME2024} \\
\cdashline{1-12}\noalign{\vskip 0.3ex}
\ensft & 47.50 & 53.33 & 43.33 & 59.17 & 50.00 & 49.17 & 39.17 & 50.00 & 48.33 & 57.50 & 49.75 \\
\nativethink & 50.00 & 43.33 & 45.00 & 55.00 & 50.83 & 33.33 & 32.50 & 51.67 & 38.33 & 43.33 & 44.33 \\
\enthink & 61.67 & 59.17 & 58.33 & 58.33 & 58.33 & 56.67 & 60.83 & 56.67 & 58.33 & 60.00 & 58.83 \\
\midrule
\multicolumn{12}{c}{AIME2025} \\
\cdashline{1-12}\noalign{\vskip 0.3ex}
\ensft & 46.67 & 46.67 & 30.00 & 51.67 & 51.67 & 38.33 & 41.67 & 48.33 & 50.00 & 43.33 & 44.83 \\
\nativethink & 46.67 & 43.33 & 45.00 & 45.00 & 45.00 & 43.33 & 35.00 & 50.00 & 43.33 & 53.33 & 45.00 \\
\enthink & 43.33 & 43.33 & 50.00 & 48.33 & 40.00 & 45.00 & 50.00 & 41.67 & 45.00 & 51.67 & 45.83 \\
\midrule
\multicolumn{12}{c}{CNMO} \\
\cdashline{1-12}\noalign{\vskip 0.3ex}
\ensft & 72.22 & 65.28 & 50.00 & 66.67 & 68.06 & 61.11 & 47.22 & 68.06 & 65.28 & 70.83 & 63.47 \\
\nativethink & 69.44 & 58.33 & 63.89 & 66.67 & 73.61 & 61.11 & 52.78 & 66.67 & 62.50 & 62.50 & 63.75 \\
\enthink & 63.89 & 68.06 & 62.50 & 72.22 & 72.22 & 68.06 & 66.67 & 70.83 & 69.44 & 66.67 & 68.06 \\
\midrule
\multicolumn{12}{c}{MATH500} \\
\cdashline{1-12}\noalign{\vskip 0.3ex}
\ensft & 95.02 & 95.10 & 85.45 & 93.09 & 94.61 & 89.55 & 88.67 & 93.33 & 88.10 & 91.64 & 91.45 \\
\nativethink & 95.42 & 94.29 & 92.44 & 94.05 & 94.61 & 89.95 & 91.16 & 93.57 & 88.99 & 93.01 & 92.75 \\
\enthink & 95.10 & 94.69 & 92.44 & 94.86 & 94.94 & 93.89 & 93.97 & 95.26 & 92.36 & 94.05 & 94.16 \\
\midrule
\multicolumn{12}{c}{MMATH} \\
\cdashline{1-12}\noalign{\vskip 0.3ex}
\ensft & 65.35 & 65.09 & 52.20 & 67.65 & 66.08 & 59.54 & 54.18 & 64.93 & 62.93 & 65.83 & 62.38 \\
\nativethink & 65.38 & 59.82 & 61.58 & 65.18 & 66.01 & 56.93 & 52.86 & 65.48 & 58.29 & 63.04 & 61.46 \\
\enthink & 66.00 & 66.31 & 65.82 & 68.44 & 66.37 & 65.90 & 67.87 & 66.11 & 66.29 & 68.10 & 66.72 \\
\bottomrule
\end{tabular}
}
\caption{Evaluation results of different training methods on Qwen2.5-32B-Instruct: \ensft (fully English fine-tuning), \nativethink (full native-language reasoning), and \enthink (English reasoning with native questions and answers).}
\label{tab:trained_results_all}
\end{table*}

\begin{figure*}[htbp]
\centering
\begin{subfigure}{0.49\linewidth}
\includegraphics[width=\linewidth]{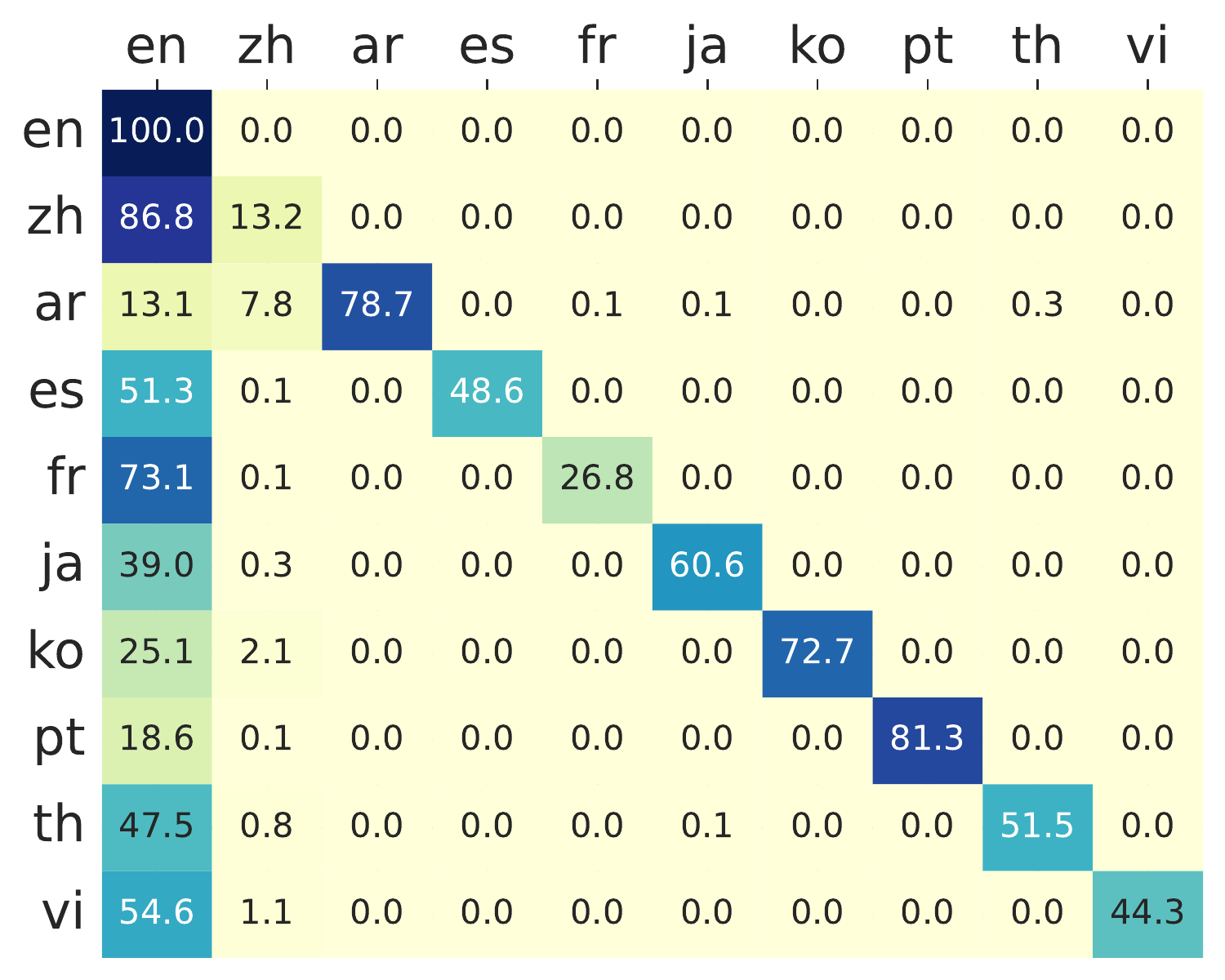}
\caption{Qwen2.5-32B-Instruct-\ensft}
\end{subfigure}
\begin{subfigure}{0.49\linewidth}
\includegraphics[width=\linewidth]{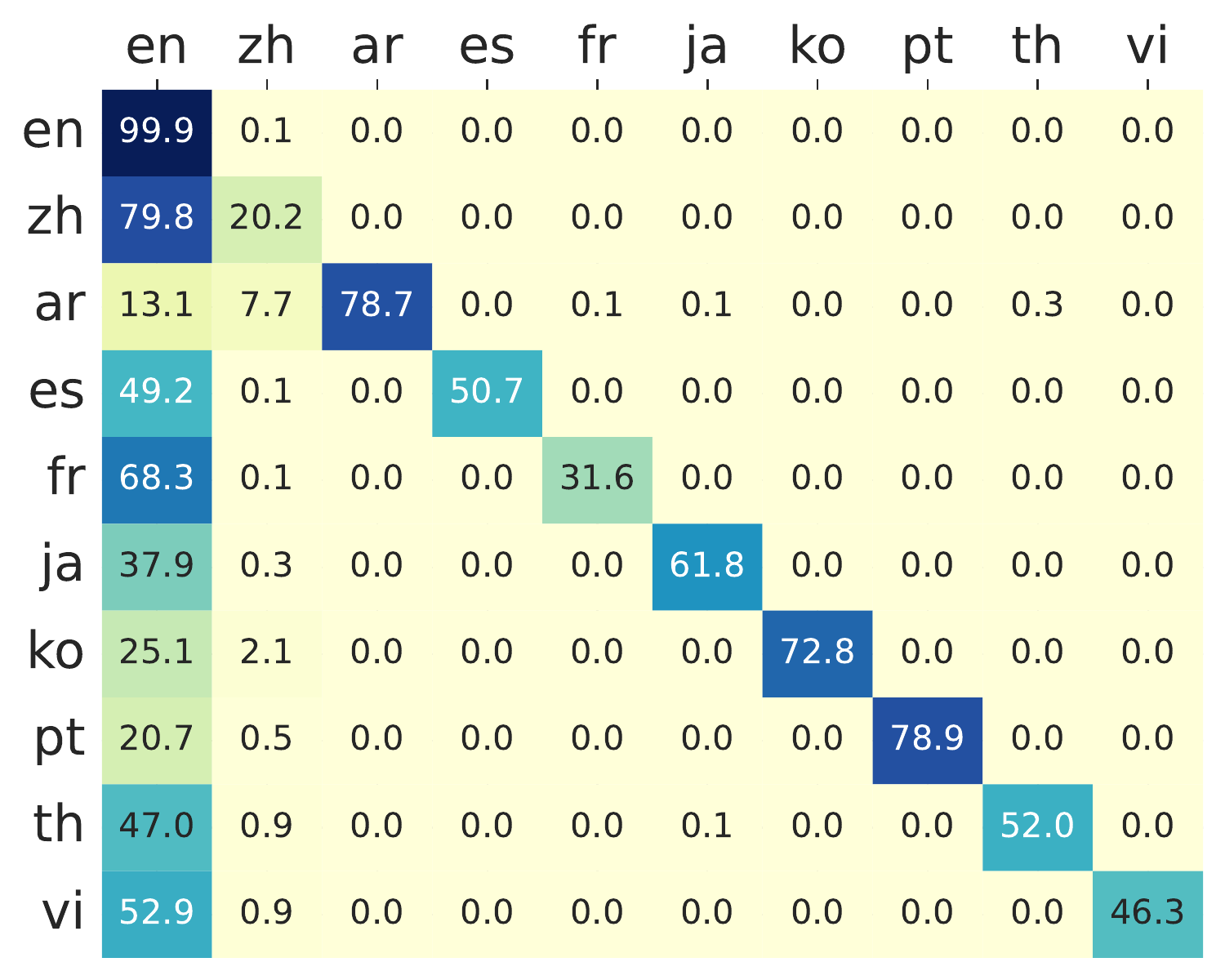}
\caption{Qwen2.5-32B-Instruct-\ensft}
\end{subfigure}

\begin{subfigure}{0.49\linewidth}
\includegraphics[width=\linewidth]{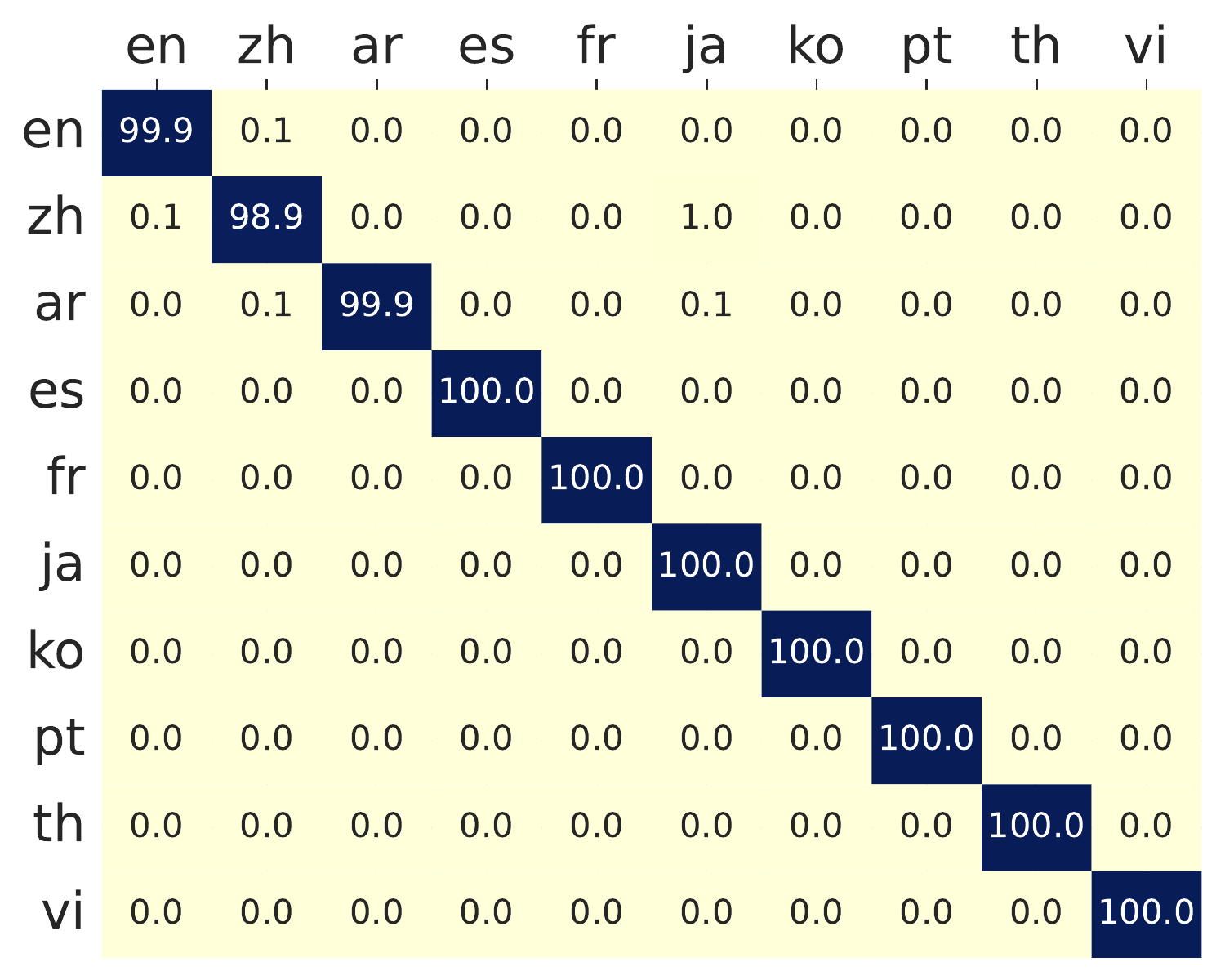}
\caption{Qwen2.5-32B-Instruct-\nativethink}
\end{subfigure}
\begin{subfigure}{0.49\linewidth}
\includegraphics[width=\linewidth]{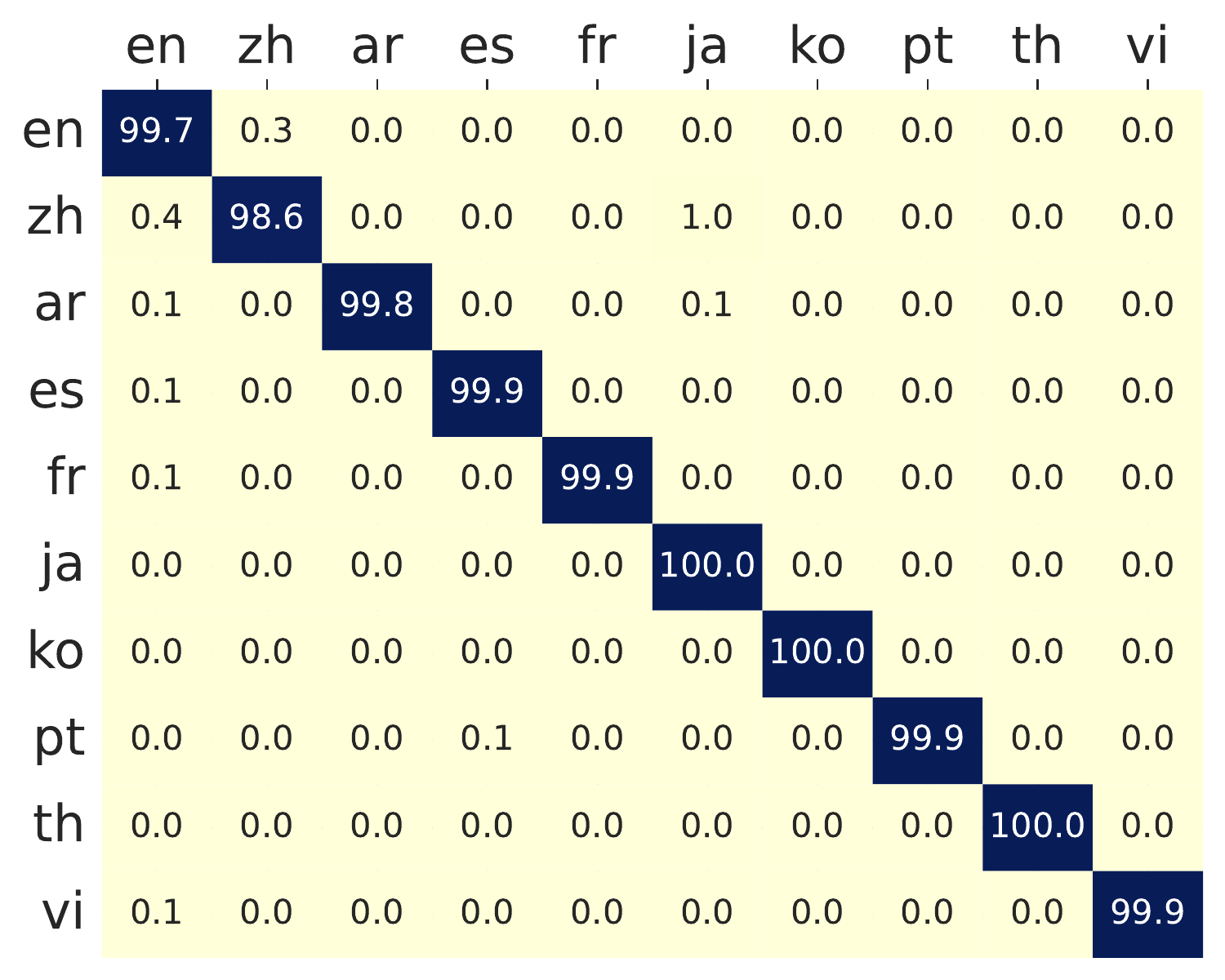}
\caption{Qwen2.5-32B-Instruct-\nativethink}
\end{subfigure}

\begin{subfigure}{0.49\linewidth}
\includegraphics[width=\linewidth]{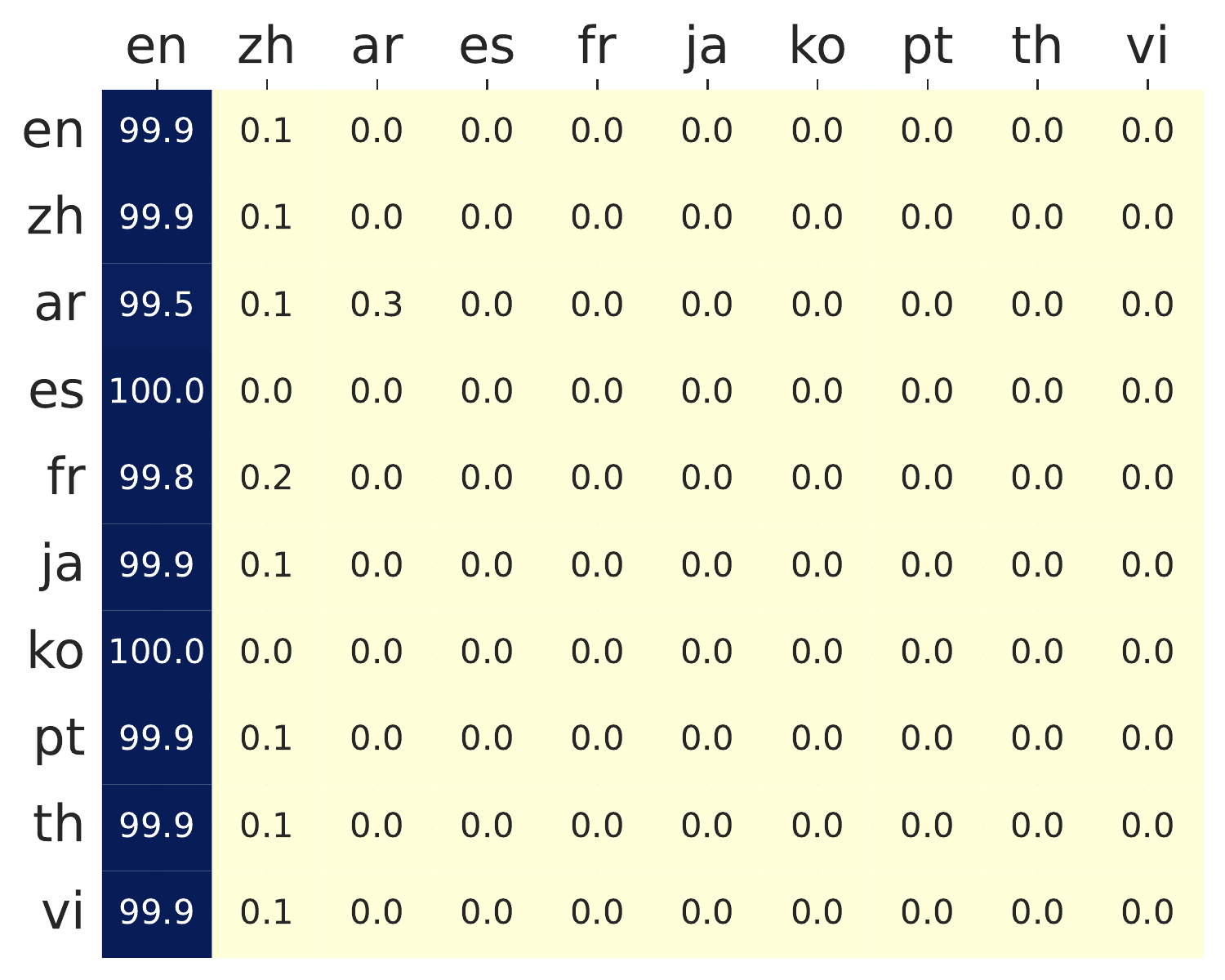}
\caption{Qwen2.5-32B-Instruct-\enthink}
\end{subfigure}
\begin{subfigure}{0.49\linewidth}
\includegraphics[width=\linewidth]{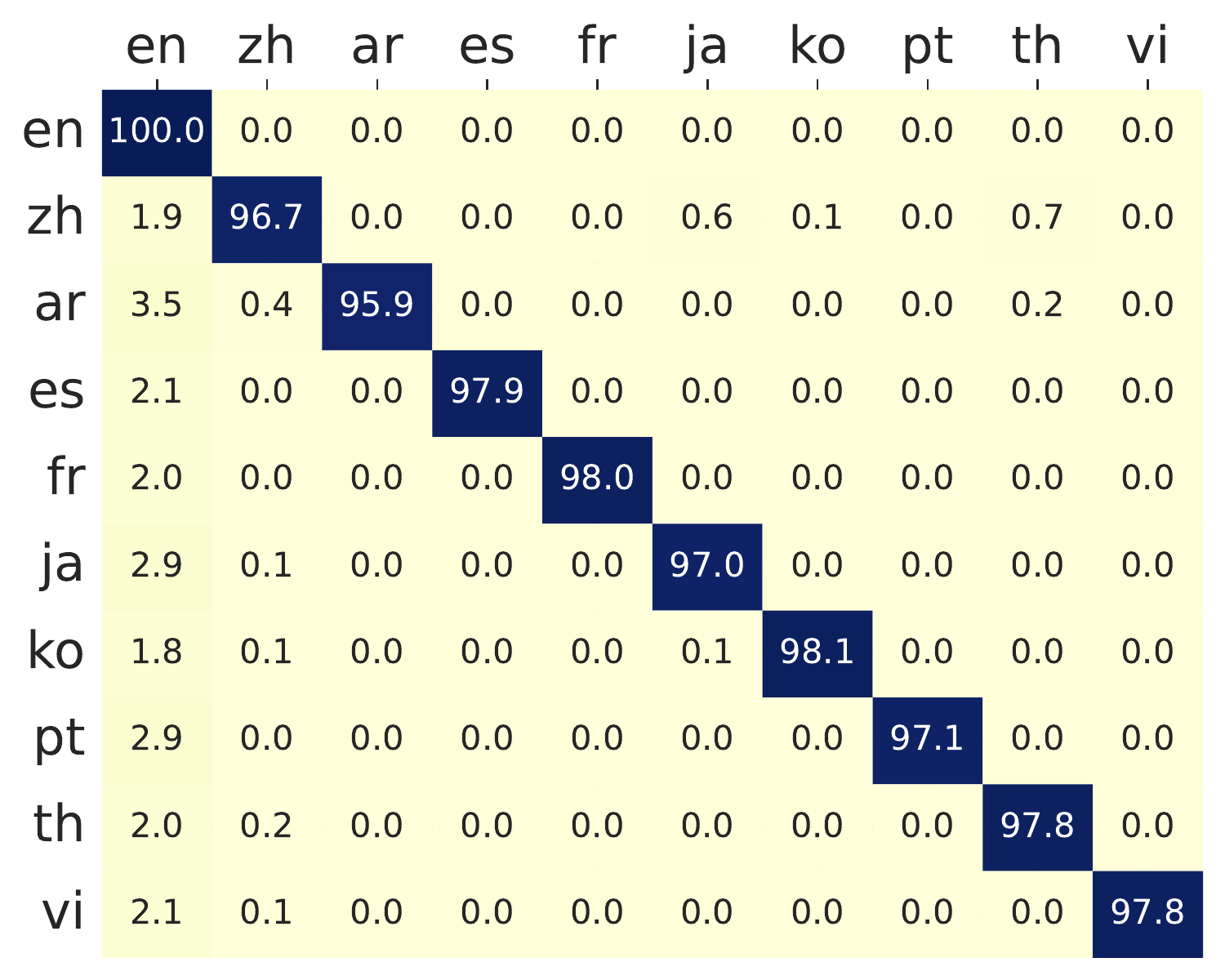}
\caption{Qwen2.5-32B-Instruct-\enthink}
\end{subfigure}

\caption{The percentage to think and answer in each language for our training methods: \ensft (fully English fine-tuning), \nativethink (full native-language reasoning), and \enthink (English reasoning with native questions and answers). The left column is the percentage of thinking, and the right column is answering.}
\label{fig:trained_results_think_ans_lagnuage}
\end{figure*}

\end{document}